\documentclass[twoside,11pt]{article}
\usepackage{nrc}
\usepackage{theapa}
\usepackage{amsmath}
\usepackage{color}
\usepackage[normalem]{ulem}
\usepackage{amssymb}            
\usepackage{enumitem}           
\usepackage{graphicx}           
\usepackage[hyphens]{url}       
\usepackage{siunitx}            
\usepackage{multirow}           
\nrcheading{2014}{May 22, 2014}
\ShortHeadings{Semantic Composition and Decomposition}{Turney}
\firstpageno{1}
\makeatletter
\newcommand\figcaption{\def\@captype{figure}\caption}
\newcommand\tabcaption{\def\@captype{table}\caption}
\makeatother
\begin{document}

\title{Semantic Composition and Decomposition: \\
       From Recognition to Generation}

\author{\name Peter D. Turney \email peter.turney@nrc-cnrc.gc.ca \\
       \addr National Research Council Canada \\
       Ottawa, Ontario, Canada, K1A 0R6 }

\maketitle

\begin{abstract}
Semantic composition is the task of understanding the meaning of text
by composing the meanings of the individual words in the text. Semantic
decomposition is the task of understanding the meaning of an individual
word by decomposing it into various aspects (factors, constituents, components)
that are latent in the meaning of the word. We take a distributional approach to
semantics, in which a word is represented by a context vector. Much recent work has
considered the problem of {\em recognizing} compositions and decompositions,
but we tackle the more difficult {\em generation} problem. For simplicity, we focus
on noun-modifier bigrams and noun unigrams. A test for semantic composition
is, given context vectors for the noun and modifier in a noun-modifier bigram
({\em red salmon}), generate a noun unigram that is synonymous with the given
bigram ({\em sockeye}). A test for semantic decomposition is, given a context
vector for a noun unigram ({\em snifter}), generate a noun-modifier bigram
that is synonymous with the given unigram ({\em brandy glass}). With a
vocabulary of about 73,000 unigrams from WordNet, there are 73,000 candidate
unigram compositions for a bigram and 5,300,000,000 (73,000 squared) candidate
bigram decompositions for a unigram. We generate ranked lists of potential
solutions in two passes. A fast unsupervised learning algorithm generates
an initial list of candidates and then a slower supervised learning algorithm
refines the list. We evaluate the candidate solutions by comparing them to WordNet
synonym sets. For decomposition (unigram to bigram), the top 100 most
highly ranked bigrams include a WordNet synonym of the given unigram 50.7\% of the
time. For composition (bigram to unigram), the top 100 most highly
ranked unigrams include a WordNet synonym of the given bigram 77.8\% of the time.
\end{abstract}

\section{Introduction}
\label{sec:intro}

Distributional semantics is based on the hypothesis that words that occur in similar
contexts tend to have similar meanings \cite{harris54,firth57}. This hypothesis
leads naturally to vector space models, in which words are represented by
context vectors \cite{turney10}. Much recent work has been concerned with the
problem of extending vector space models beyond words, to phrases and sentences
\cite{erk13}. The approach that has worked with words does not scale up
to phrases and sentences, due to data sparsity and growth in model size \cite{turney12}.

In this paper, we focus on noun-modifier bigrams as a relatively simple testbed
for exploring the problem of modeling phrases. A noun-modifier bigram ({\em milk sugar})
consists of a head noun ({\em sugar}) and a noun or adjective ({\em milk}) that modifies
the meaning of the head noun. Given context vectors for the noun and the
modifier, we would like to model the meaning of the noun-modifier bigram.
A natural test of a model is whether it can recognize when a noun unigram
({\em lactose}) is synonymous with a bigram ({\em milk sugar}). Given a
target bigram and a list of candidate unigrams, can the model recognize
the synonymous unigram among the candidates \cite{turney12}? Given a
target unigram and a list of candidate bigrams, can the model recognize
the synonymous bigram among the candidates \cite{kartsaklis12}? 

Many noun unigrams have synonymous noun-modifier bigrams. For example, the WordNet
synonym set for {\em contrabass} is {\em bass fiddle, bass viol, bull fiddle, double
bass, contrabass}, and {\em string bass}.\footnote{WordNet is available at
\url{http://wordnet.princeton.edu/}. Our experiments use WordNet 3.0 for Linux.}
Each of the bigrams may be viewed as a {\em decomposition} of the unigram
{\em contrabass} into parts (factors, aspects, constituents). For example, the bigram
{\em bass fiddle} expresses that a {\em contrabass} is a type of {\em fiddle} that
covers the {\em bass} range; the head noun {\em fiddle} expresses the {\em type}
aspect of {\em contrabass} and the modifier {\em bass} expresses the {\em range}
aspect of {\em contrabass}. Decomposition (unigram $\rightarrow$ bigram) maps a word
({\em contrabass}) to a (miniature) definition ({\em bass fiddle}), whereas
composition (bigram $\rightarrow$ unigram) maps a definition to a word (by composing
the components of the definition into a whole).

Past work has achieved promising results on recognizing noun-modifier compositions
and decompositions, given relatively short lists of candidates. For example,
\citeA{turney13} achieves 75.9\% accuracy on noun-modifier composition recognition,
given seven candidate unigrams for each target bigram, and \citeA{kartsaklis12}
achieve 24\% accuracy on noun decomposition recognition, given seventy-two choices.
However, the degree of challenge in the recognition task depends on the nature of the
lists of candidates. If the distractors are very different from the correct choice, then
the task is easy. The generation task avoids this criticism, because there is no
predetermined list of choices. The choices are only limited by the vocabulary.

In this work, our main resources are the WordNet lexicon and the Waterloo corpus.
The corpus consists of web pages gathered from university websites with a webcrawler,
by Charles Clarke at the University of Waterloo. The corpus contains about $5 \times 10^{10}$
words. Stored as plain text, without HTML markup, the corpus size is 280 gigabytes.

From the terms in WordNet, we extracted approximately 114,000 $n$-grams that occur
at least 100 times in the Waterloo corpus. The $n$-grams include about 73,000
unigrams, 36,000 bigrams, and 5,000 $n$-grams with $n > 2$. The majority of the
unigrams are nouns and the majority of the bigrams are noun-modifier phrases.\footnote{We
do not attempt to filter out the unigrams that are not nouns and the bigrams that
are not noun-modifiers, although it would be easy to do so, using the information in
WordNet. The intent is that the algorithms should be robust enough to handle them
automatically, using only corpus-based information.}

For composition (bigram $\rightarrow$ unigram), given a target bigram ({\em bass fiddle}),
we use a fast unsupervised algorithm, Comp, to score the 73,000 unigrams from WordNet,
and we output the top 2000 highest scoring candidates. We then use a slower supervised
algorithm, Super, to rescore the top 2000 candidates from Comp. Our aim is to have
a synonymous unigram ({\em contrabass}) somewhere in the top 100 rescored candidates
from Super. We achieve this for 77.8\% of the 351 target bigrams in our test dataset.

For decomposition (unigram $\rightarrow$ bigram), given a target unigram ({\em contrabass}),
we use a fast unsupervised algorithm, Decomp, to score the 73,000 unigrams from WordNet.
First the unigrams are scored as candidate modifiers and then they are scored as
candidate heads, resulting in a list of the top 1000 potential modifiers and another list
of the top 1000 potential heads. Combining these two lists by concatenation of every
top modifier with every top head, we form a list of 1,000,000 candidate bigrams. These
bigrams are then scored and we output the top 2000 highest scoring candidates. We then use
the slower supervised algorithm, Super, to rescore the top 2000 candidates from Decomp.
50.7\% of the time, a synonymous bigram ({\em bass fiddle}) is among the top 100 rescored
candidates for the 355 target unigrams in our test dataset.

Comp and Decomp are variations on the unsupervised learning algorithm of \citeA{turney12}.
Super is based on the supervised algorithm of \citeA{turney13}. These algorithms were
originally designed for the recognition task. The main contribution of this paper is to
show that, working together, the algorithms can scale up from recognition to generation.

In Section~\ref{sec:related}, we review related work on recognizing compositions
and decompositions. The datasets we use for evaluating the algorithms are presented
in Section~\ref{sec:datasets}. The three algorithms (Comp, Decomp, and Super)
share a set of features, which we describe in Section~\ref{sec:features}. The algorithms
are presented in Section~\ref{sec:algorithms}. Experiments with composition 
are given in Section~\ref{sec:composition} and Section~\ref{sec:decomposition} covers the
experiments with decomposition. We summarize the results in Section~\ref{sec:discussion}.
Section~\ref{sec:future} discusses limitations of this work and we conclude
in Section~\ref{sec:conclusions}.

\section{Related Work}
\label{sec:related}

We first examine related work on paraphrase in general and then look at work on
composition and decomposition.

\subsection{Paraphrasing}
\label{subsec:rel-para}

Mapping between nouns and noun-modifiers is a form of paraphrasing. \citeA{madnani10} and
\citeA{androutsopoulos10} present thorough surveys of data-driven approaches to paraphrasing.
Paraphrases can also be generated using knowledge-based techniques, but our focus here is
corpus-based methods, since they generally require much less human effort than 
knowledge-based techniques.

In general, corpus-based approaches to paraphrase have extended the distributional hypothesis
from words to phrases. The extended distributional hypothesis is that {\em phrases} that occur in
similar contexts tend to have similar meanings \cite{lin01}. For example, consider the following
fragments of text \cite{pasca05}:

\begin{itemize}[itemsep=1pt,parsep=1pt,topsep=4pt,partopsep=1pt]    

\item {\em 1989, when Soviet troops} withdrew from {\em Afghanistan}

\item {\em 1989, when Soviet troops} pulled out of {\em Afghanistan}

\end{itemize}

\noindent From the shared context, we can infer a degree of semantic similarity
between the phrases {\em withdrew from} and {\em pulled out of}. We call this the {\em holistic}
(non-compositional) approach to paraphrase \cite{turney12}, because the phrases are treated as
opaque wholes. The holistic approach does not model the individual words in the phrases.

The creative power of language comes from combining words to create new meanings.
With a vocabulary of $N$ unigrams, there are $N^2$ possible bigrams and $N^3$ possible trigrams.
We give meaning to $n$-grams ($n > 1$) by composing the meanings of their component words.
The holistic approach lacks the ability to compose meanings and cannot scale up
to phrases and sentences. Holistic approaches to paraphrase do not address the creative
power of language \cite{chomsky75,fodor02}.

The holistic approach often achieves excellent results. It is especially suited to idiomatic
phrases ({\em kick the bucket}), but, eventually, as we consider longer and more diverse phrases,
we encounter problems with data sparsity and model size. In this paper, we concentrate on
compositional models, but we include holistic models as baselines in our experiments. 

\subsection{Composition}
\label{subsec:rel-comp}

Let $ab$ be a noun-modifier phrase, and assume that we have context vectors $\mathbf{a}$ and
$\mathbf{b}$ that represent the component words $a$ and $b$. One of the earliest proposals for
semantic composition is to represent the bigram $ab$ by the vector sum $\mathbf{a} + \mathbf{b}$
\cite{landauer97}. To measure the similarity of a noun-modifier phrase, $ab$, and a noun, $c$,
we calculate the cosine of the angle between $\mathbf{a} + \mathbf{b}$ and the context vector
$\mathbf{c}$ for $c$.

This simple proposal actually works relatively well \cite{mitchell08,mitchell10},
although it lacks order sensitivity. Since $\mathbf{a} + \mathbf{b} = \mathbf{b} + \mathbf{a}$,
{\em animal farm} and {\em farm animal} have the same representation, although one
is a type of farm and the other is a type of animal. \citeA{landauer02} estimates that 80\% of
the meaning of English text comes from word choice and the remaining 20\% comes from
word order, thus vector addition misses at least 20\% of the meaning of a bigram.

\citeA{kintsch01} and \citeA{utsumi09} propose variations of additive composition in which
$ab$ is represented by $\mathbf{a} + \mathbf{b} + \sum_i \mathbf{n}_i$, the sum of $\mathbf{a}$,
$\mathbf{b}$, and selected neighbours $\mathbf{n}_i$ of $\mathbf{a}$ and $\mathbf{b}$.
The neighbours are context vectors for other words in the given vocabulary. \citeA{mitchell10}
found that a simple additive model peformed better than an additive model that included neighbours.

\citeA{mitchell08,mitchell10} suggest element-wise multiplication as a composition operation,
$\mathbf{c} = \mathbf{a} \odot \mathbf{b}$, where $c_i = a_i \, \cdot \, b_i$. Since
$\mathbf{a} \odot \mathbf{b} = \mathbf{b} \odot \mathbf{a}$, element-wise multiplication
is not sensitive to word order. However, in an experimental evaluation of seven compositional
models and two noncompositional models, element-wise multiplication had the best performance
\cite{mitchell10}.

In the holistic approach, $ab$ is treated as if it were an individual word. A context vector
for $ab$ is constructed from a corpus in the same manner as it would be constructed for a unigram.
This approach does not scale up, but it does work well for a predetermined small set
of high frequency $n$-grams \cite{turney12}. \citeA{guevara10} and \citeA{baroni10} point out that
a small set of bigrams with holistic context vectors can be used to train a regression model.
For example, a regression model can be trained to map the context vectors $\mathbf{a}$ and
$\mathbf{b}$ to the holistic context vector for $ab$ \cite{guevara10}. Given a new bigram, $cd$,
with context vectors $\mathbf{c}$ and $\mathbf{d}$, the regression model can use
$\mathbf{c}$ and $\mathbf{d}$ to predict the holistic context vector for $cd$.

Many other ideas have been proposed for extending distributional semantics to phrases and
sentences. Recently there have been several overviews of this topic \cite{mitchell10,turney12,erk13}.
Most of the proposed extensions to distributional semantics involve operations from linear algebra,
such as tensor products \cite{clark07,widdows08,clark08,grefenstette11}. Another proposal is to
operate on similarities instead of (or in addition to) working directly with context vectors
\cite{socher11,turney12,turney13}.

For example, let ${\rm sim_s}$ be a measure of semantic similarity for words, phrases, and sentences.
Let ${\rm sim_v}$ be a measure of similarity (or, conversely, distance) for vectors, matrices, and tensors,
such as cosine, Euclidean distance, inner product, or Frobenius norm. Given a bigram $ab$, a unigram
$c$, and context vectors, $\mathbf{a}$, $\mathbf{b}$, and $\mathbf{c}$, suppose that we are seeking
a good measure of the semantic similarity between $ab$ and $c$, ${\rm sim_s}(ab,c)$. This semantic
similarity must somehow compose $a$ and $b$ in order to recognize the similarity between $ab$ and $c$.

One approach to capturing ${\rm sim_s}(ab,c)$ is to search for a function, $\mathbf{f}$, that composes
vectors (or matrices or tensors), as follows:

\begin{equation}
{\rm sim_s}(ab,c) = {\rm sim_v}(\mathbf{f}(\mathbf{a},\mathbf{b}),\mathbf{c})
\end{equation}

\noindent Element-wise multiplication \cite{mitchell08,mitchell10} is an instance of this approach,
where $\mathbf{f}(\mathbf{a},\mathbf{b}) = \mathbf{a} \odot \mathbf{b}$, as follows:

\begin{equation}
{\rm sim_s}(ab,c) = {\rm cos}(\mathbf{a} \odot \mathbf{b},\mathbf{c})
\end{equation}

\noindent Much work focuses on finding the right $\mathbf{f}$ for various types of semantic
composition \cite{clark07,widdows08,mitchell08,mitchell10,guevara10,grefenstette11}. We call
this general approach {\em context composition}, due to the arguments of the function $\mathbf{f}$.

Another approach attempts to capture ${\rm sim_s}(ab,c)$ by searching for a function, $f$, that
composes similarities, as follows:

\begin{equation}
{\rm sim_s}(ab,c) = f({\rm sim_v}(\mathbf{a},\mathbf{c}), {\rm sim_v}(\mathbf{b},\mathbf{c}))
\end{equation}

\noindent For instance, $f$ could be the geometric mean (see Equation~\ref{eq:scoreu} in
Section~\ref{subsec:comp}), as follows:

\begin{equation}
{\rm sim_s}(ab,c) = \sqrt{{\rm cos}(\mathbf{a},\mathbf{c}) \cdot {\rm cos}(\mathbf{b},\mathbf{c})}
\end{equation}

\noindent \citeA{turney12} took the second approach to recognizing compositions, using hand-built
functions for $f$. \citeA{turney13} used supervised learning to find $f$. We call this general
approach {\em similarity composition}, based on the arguments of $f$.

\citeA{socher11} combined context composition and similarity composition for sentence paraphrase
recognition. Unsupervised recursive autoencoders were used to compose context vectors and then a
supervised softmax classifier was used to compose a similarity matrix.

\citeA{turney12} introduced a dataset of 2180 semantic composition questions, split
into 680 questions for training and 1500 for testing.\footnote{The dataset is available at
\url{http://jair.org/papers/paper3640.html}.} Table~\ref{tab:turney12} shows one of the questions.
The stem is the target noun-modifier bigram and there are seven candidate unigrams.
These questions were generated automatically from WordNet. The stem and the solution
always belong to the same WordNet synonym set. The intention is to evaluate a proposed similarity
measure, ${\rm sim_s}(ab,c)$, by its accuracy on the 1500 testing questions.

\begin{table}
\begin{center}
\scalebox{0.9}{
\begin{tabular}{lll}
\hline
Stem:      &       & heart disease \\
\hline
Choices:   & (1)   & cardiopathy \\
           & (2)   & heart \\
           & (3)   & disease \\
           & (4)   & bosom \\
           & (5)   & illness \\
           & (6)   & gimel \\
           & (7)   & betatron \\
\hline
Solution:  & (1)   & cardiopathy \\
\hline
\end{tabular}
} 
\end{center}
\caption {A seven-choice noun-modifier question based on WordNet \cite{turney12}.}
\label{tab:turney12}
\end{table}

Using an unsupervised learning algorithm and a hand-built function for $f$, \citeA{turney12}
achieved an accuracy of 58.3\% on the 1500 testing questions. Vector addition reached 50.1\%
and element-wise multiplication attained 57.5\%. \citeA{turney13} used supervised learning to
optimize $f$ on the training questions, yielding an accuracy of 75.9\% on the testing questions.
The holistic approach gets 81.6\% correct, but suffers from scaling problems.

\citeA{dinu13} created a dataset of 620 adjective-noun phrases, each paired with a noun unigram
from the same WordNet synonym set. Table~\ref{tab:dinu13} shows the first four phrases in their
dataset. Given a vocabulary of 21,000 noun unigrams, the task of each distributional model
was to rank the nouns by their semantic similarity to each adjective-noun phrase.

\begin{table}
\begin{center}
\scalebox{0.9}{
\begin{tabular}{lll}
\hline
Target               & $\rightarrow$   & Solution \\
\hline
double star          & $\rightarrow$   & binary \\
monosyllabic word    & $\rightarrow$   & monosyllable \\
presidential term    & $\rightarrow$   & presidency \\
electrical power     & $\rightarrow$   & wattage \\
\hline
\end{tabular}
} 
\end{center}
\caption {Adjective-noun composition problems based on WordNet \cite{dinu13}.}
\label{tab:dinu13}
\end{table}

\citeA{dinu13} evaluated seven different models on their adjective-noun dataset. The performance
of the models was measured by the medians of the ranks of the solution nouns in the ranked lists of 21,000
candidates. Our experiments (Sections \ref{sec:composition} and \ref{sec:decomposition}) use the same
general approach to evaluation of models.

In a noun-modifier phrase, the modifier may be either a noun or an adjective; therefore
adjective-noun phrases are a subset of noun-modifier phrases. \citeA{dinu13} hypothesize that
adjectives are functions that map nouns onto modified nouns \cite{baroni10}, thus they
believe that noun-noun phrases and adjective-noun phrases should have different kinds of
models. The models we present here (Section~\ref{sec:algorithms}) treat all noun-modifiers
the same way, hence our datasets contain both noun-noun phrases and adjective-noun phrases.
For comparison, we will also evaluate our models on \citeS{dinu13} adjective-noun dataset
(Section~\ref{subsec:comp-adj-noun}).

\subsection{Decomposition}
\label{subsec:rel-decomp}

\citeA{collins-thompson07} describe the {\em definition production task}, in which
human subjects are asked to generate a short definition of a given target word,
to evaluate their understanding of the word. They propose an algorithm for automatically
scoring the answers of the human subjects. Their algorithm measures the semantic
similarity between a gold standard reference definition, given by a human expert, and
the human subject's definition. The semantic similarity is calculated using a Markov chain.

\citeA{kartsaklis12} created a dataset with 72 target noun unigrams and 40 target verb
unigrams.\footnote{The dataset is available at
\url{http://www.cs.ox.ac.uk/activities/compdistmeaning/index.html}.}
Each target unigram has three gold standard definitions, where the definitions contain
2.4 words on average (the majority are bigrams). Table~\ref{tab:kartsaklis12} shows the
first four target terms and their corresponding definitions. 

\begin{table}
\begin{center}
\scalebox{0.9}{
\begin{tabular}{ll}
\hline
Target term     & Definitions \\
\hline
alley           & narrow street, small road, tiny passage \\
anorak          & waterproof jacket, rainproof coat, nonabsorbent cover \\
boy             & male child, masculine youngster, manlike kid \\
baby            & young child, infant person, new kid \\
\hline
\end{tabular}
} 
\end{center}
\caption {A sample of the dataset for the term-definition task \cite{kartsaklis12}.}
\label{tab:kartsaklis12}
\end{table}

\citeA{kartsaklis12} treat each target term (unigram) as a class label and evaluate each model
by its accuracy on classifying the definitions. For the noun unigrams, there are 216
(72 $\times$ 3) definitions, and each definition must be assigned to one of 72 classes
(target nouns). For the verb unigrams, there are 120 (40 $\times$ 3) definitions and
40 classes. Their model, based on matrix multiplication combined with element-wise multiplication,
achieves an accuracy of 24\% on the nouns and 28\% on the verbs. Element-wise multiplication
on its own attains an accuracy of 22\% on the nouns and 30\% on the verbs.

Recently \citeA{dinu14} partially addressed the task of decomposing a noun unigram into an
adjective-noun bigram. For a given noun, they generated a ranked list of candidate adjectives
and a ranked list of candidate nouns, but they did not attempt to rank the combined
adjective-noun bigrams.

\section{Datasets}
\label{sec:datasets}

In this section, we introduce the four datasets that will be used in the experiments. The
two {\em standard} datasets are based on WordNet synonym sets. The two {\em holistic}
datasets are based on the holistic approach to bigrams. All four datasets were created
with the WordNet::QueryData Perl package.\footnote{The four datasets are available on
request from the author.}

The two standard datasets are derived from the 2180 seven-choice noun-modifier questions
used in previous work \cite{turney12} to test composition recognition (see Table~\ref{tab:turney12}
in Section~\ref{subsec:rel-comp}). The original dataset included both compositional
bigrams ({\em magnetic force}) and idiomatic (non-compositional) bigrams ({\em stool pigeon}).
Given the difficulty of the generation problem, relative to the recognition problem, we decided
to make the dataset easier by avoiding idiomatic bigrams. We used WordNet glosses as heuristic
clues for finding compositional bigrams. A bigram was considered to be {\em highly compositional}
if there was at least one gloss that contained the head noun and at least one gloss (not necessarily
the same gloss; possibly a gloss for another sense of the bigram) that contained the modifier.
To allow for derivationally related word forms, a gloss was deemed to contain a word if the first
five characters of the word matched the first five characters of any word in the gloss.

For example, {\em stool pigeon} has the glosses {\em someone acting as an informer or decoy for
the police} and {\em a dummy pigeon used to decoy others}. The head noun {\em pigeon} occurs
in the second gloss, but the modifier {\em stool} appears in neither gloss. Therefore
{\em stool pigeon} fails the test; it is {\em not} considered to be highly compositional.
On the other hand, {\em magnetic force} has the gloss {\em attraction for iron; associated with
electric currents as well as magnets; characterized by fields of force}. The head noun
{\em force} occurs in the gloss and the first five characters of the modifier {\em magnetic}
match the first five characters of the word {\em magnets} in the gloss. Therefore 
{\em magnetic force} is deemed to be highly compositional.

The {\em standard composition dataset} (bigram $\rightarrow$ unigram) was generated by
extracting the stem bigram from each of the 2180 seven-choice noun-modifier questions
and checking the glosses of each bigram to see whether the bigram is highly compositional. Bigrams
that passed the test were selected as target bigrams for the standard composition dataset.
For each target bigram, all of the unigrams that are in the same WordNet synonym set as the bigram
were selected as solutions for that bigram. The standard composition dataset was divided into
training and testing subsets, based on whether the target bigram came from the training or
testing subset of the original 2180 seven-choice noun-modifier questions.

The {\em standard decomposition dataset} (unigram $\rightarrow$ bigram) was generated by
extracting the solution unigram from each of the 2180 seven-choice noun-modifier questions.
For each unigram, all of the bigrams that are in the same WordNet synonym set as the unigram
were selected as solutions for that unigram. The solution bigrams were then checked to see
whether at least one of the solution bigrams was highly compositional. If the unigram had
one highly compositional solution, then the unigram was selected as a target unigram for
the standard decomposition dataset, and all of the bigrams in the same WordNet synonym
set as the unigram were selected as solutions for that unigram. The standard decomposition dataset
was divided into training and testing subsets, based on whether the target unigram came from the
training or testing subset of the original 2180 seven-choice noun-modifier questions.

The two standard datasets exploit the human expertise and effort that went into the construction of
WordNet synonym sets. The idea of the holistic datasets is to reduce the need for human expertise,
by using holistic vectors instead of synonym sets. This idea was inspired by \citeA{guevara10}
and \citeA{baroni10}, who used holistic vectors to train regression models. 

Let's use $a\_b$ to represent a bigram that we will represent with a holistic vector.
That is, although $a\_b$ is a bigram, we will pretend it is a unigram and we will
construct a context vector for it in the same manner as we would for any unigram.
We call $a\_b$ a {\em pseudo-unigram}. The {\em holistic composition dataset}
consists of true bigram targets and pseudo-unigram solutions. The {\em holistic
decomposition dataset} consists of pseudo-unigram targets and true bigram solutions.

The idea is that mapping {\em red salmon} to {\em sockeye} (standard composition)
is analogous to mapping {\em red salmon} to {\em red\_salmon} (holistic composition),
and mapping {\em sockeye} to {\em red salmon} (standard decomposition) is
analogous to mapping {\em red\_salmon} to {\em red salmon} (holistic decomposition).
Although the mapping between  {\em red\_salmon} and {\em red salmon} is trivial
when we have access to their written forms, it is not trivial when we have only
the three corresponding context vectors (for {\em red}, {\em salmon}, and {\em red\_salmon}).

The two holistic datasets are based on bigrams selected from the 36,000 bigrams
in WordNet. We calculated the frequency of each bigram in our corpus and used the
most frequent bigrams to construct targets and solutions. The sizes of the holistic
datasets were matched to the sizes of the corresponding standard datasets.

For each of the four datasets, Table~\ref{tab:datasets-samples} shows the first
four targets and their solutions. Table~\ref{tab:datasets} gives the sizes of
datasets. Each holistic target has only one solution but the standard targets
may have more than one solution. When there are several possible solutions, an
algorithm is considered successful if a guess matches any of the possible soutions.

\begin{table}
\begin{center}
\scalebox{0.9}{
\begin{tabular}{lllcl}
\hline
Dataset       & & Target              & $\rightarrow$ & Solutions \\
\hline
              & \multirow{4}{*}{$\left\{\rule{0pt}{30pt}\right.$}
                & red salmon          & $\rightarrow$ & sockeye \\
standard      & & brandy glass        & $\rightarrow$ & snifter \\
composition   & & past perfect        & $\rightarrow$ & pluperfect \\
              & & foot lever          & $\rightarrow$ & pedal, treadle \\[4pt]
              & \multirow{5}{*}{$\left\{\rule{0pt}{37pt}\right.$}
                & sockeye             & $\rightarrow$ & blueback salmon, red salmon, \\
standard      & &                     &               & sockeye salmon, oncorhynchus nerka \\
decomposition & & snifter             & $\rightarrow$ & brandy snifter, brandy glass \\
              & & pluperfect          & $\rightarrow$ & past perfect, pluperfect tense \\
              & & treadle             & $\rightarrow$ & foot pedal, foot lever \\[4pt]
              & \multirow{4}{*}{$\left\{\rule{0pt}{30pt}\right.$}
                & art history         & $\rightarrow$ & art\_history \\
holistic      & & fast track          & $\rightarrow$ & fast\_track \\
composition   & & good deal           & $\rightarrow$ & good\_deal \\
              & & standing committee  & $\rightarrow$ & standing\_committee \\[4pt]
              & \multirow{4}{*}{$\left\{\rule{0pt}{30pt}\right.$}
                & major\_league       & $\rightarrow$ & major league \\
holistic      & & jet\_stream         & $\rightarrow$ & jet stream \\
decomposition & & purchase\_price     & $\rightarrow$ & purchase price \\
              & & chat\_room          & $\rightarrow$ & chat room \\[2pt]
\hline
\end{tabular}
} 
\end{center}
\caption {A sample of the first four targets in each of the testing datasets.}
\label{tab:datasets-samples}
\end{table}

\begin{table}
\begin{center}
\scalebox{0.9}{
\begin{tabular}{llllccc}
\hline
Dataset       & & Target $\rightarrow$ solutions        & Split     & Number   & Number      & Ratio \\
              & &                                       &           & targets  & solutions   & sol/tar \\
\hline
standard      & \multirow{2}{*}{$\left\{\rule{0pt}{15pt}\right.$}
                & \multirow{2}{*}{bigram $\rightarrow$ unigram}
                                                        & training  & 154      & 234         & 1.5 \\
composition   & &                                       & testing   & 351      & 581         & 1.7 \\[4pt]
standard      & \multirow{2}{*}{$\left\{\rule{0pt}{15pt}\right.$}
                & \multirow{2}{*}{unigram $\rightarrow$ bigram}
                                                        & training  & 198      & 408         & 2.1 \\
decomposition & &                                       & testing   & 355      & 714         & 2.0 \\[4pt]
holistic      & \multirow{2}{*}{$\left\{\rule{0pt}{15pt}\right.$}
                & \multirow{2}{*}{bigram  $\rightarrow$ pseudo-unigram}
                                                        & training  & 154      & 154         & 1.0 \\
composition   & &                                       & testing   & 351      & 351         & 1.0 \\[4pt]
holistic      & \multirow{2}{*}{$\left\{\rule{0pt}{15pt}\right.$}
                & \multirow{2}{*}{pseudo-unigram $\rightarrow$ bigram}
                                                        & training  & 198      & 198         & 1.0 \\
decomposition & &                                       & testing   & 355      & 355         & 1.0 \\[2pt]
\hline
\end{tabular}
} 
\end{center}
\caption {Characteristics of the four composition and decomposition datasets.}
\label{tab:datasets}
\end{table}

\section{Features}
\label{sec:features}

Comp, Decomp, and Super use five types of features for ranking candidates.
The features are functions that take unigrams or pseudo-unigrams as arguments and
return real values. Tables \ref{tab:feats} and \ref{tab:vars} list the five functions
and their arguments. The features DS and FS were introduced in \citeA{turney12}.
These two features were supplemented with LUF and PPMI in \citeA{turney13}. The feature
LBF is a new addition to the group.

\begin{table}
\begin{center}
\scalebox{0.9}{
\begin{tabular}{ll}
\hline
Feature                       & Description \\
\hline
$\textrm{LUF}(a)$             & log unigram frequency \\
$\textrm{LBF}(a, b)$          & log bigram frequency \\
$\textrm{PPMI}(a, b, h)$      & positive pointwise mutual information \\
$\textrm{DS}(a, b, k, p)$     & domain similarity \\
$\textrm{FS}(a, b, k, p)$     & function similarity \\
\hline
\end{tabular}
} 
\end{center}
\caption {The five types of features.}
\label{tab:feats}
\end{table}

\begin{table}
\begin{center}
\scalebox{0.9}{
\begin{tabular}{cl}
\hline
Variable     & Description \\
\hline
$a, b$       & unigrams or pseudo-unigrams \\
$h$          & the position of $b$, {\em left} of $a$ or {\em right} of $a$, for PPMI ($h$ for {\em handedness}) \\
$k$          & the number of latent factors for the similarity measures, for DS and FS \\
$p$          & a number specifying the {\em power} of the similarity measures, for DS and FS \\
\hline
\end{tabular}
} 
\end{center}
\caption {The variables for the features.}
\label{tab:vars}
\end{table}

\subsection{LUF: Log Unigram Frequency }
\label{subsec:luf}

Let $\textrm{UF}(a)$ be the frequency of the unigram $a$ in the Waterloo corpus. We define $\textrm{LUF}(a)$
as $\textrm{log}(\textrm{UF}(a) + 1)$. We add one to the frequency because $\textrm{log}(0)$ is undefined.
If $a$ is not in the Waterloo corpus, $\textrm{UF}(a)$ is zero, and thus $\textrm{LUF}(a)$ is also zero.
For a pseudo-unigram, $a\_b$, $\textrm{UF}(a\_b)$ is the frequency of the bigram $ab$ in the Waterloo corpus.

We precompute $\textrm{UF}(a)$ for the 73,000 unigrams and 36,000 bigrams in WordNet, hence $\textrm{LUF}(a)$
covers 73,000 unigrams and 36,000 pseudo-unigrams. In our experiments, we use a hash table to quickly obtain
$\textrm{LUF}(a)$.\footnote{The frequencies of all WordNet $n$-grams in the Waterloo corpus are available on
request from the author.}

\subsection{LBF: Log Bigram Frequency}
\label{subsec:lbf}

For $\textrm{LBF}(a, b)$, we use the Google Web 1T 5-gram (Web1T5) dataset \cite{brants06}.\footnote{The
Web1T5 dataset is available at \url{http://catalog.ldc.upenn.edu/LDC2006T13}.} Let $\textrm{BF}(a, b)$
be the frequency of the bigram $ab$ in the Web1T5 dataset. We define $\textrm{LBF}(a, b)$ as
$\textrm{log}(\textrm{BF}(a, b) + 1)$. For a pseudo-unigram, $a\_b$, $\textrm{BF}(a\_b, c)$ is the
frequency of the trigram $abc$ in the Web1T5 dataset and $\textrm{BF}(c, a\_b)$ is the frequency of the
trigram $cab$. In our experiments, we never need to compute $\textrm{BF}$ when both arguments are
pseudo-unigrams, $\textrm{BF}(a\_b, c\_d)$, so we do not need to work with 4-grams.

We precompute $\textrm{BF}(a, b)$ for all bigrams and trigrams such that the component unigrams are in WordNet
and the bigrams and trigrams are in the Web1T5 dataset. This is too much data to fit in RAM, so we store the
data in a Berkeley DB database, in order to rapidly find $\textrm{LBF}(a, b)$.\footnote{We use the BerkeleyDB
Perl package.}

\subsection{PPMI: Positive Pointwise Mutual Information}
\label{subsec:ppmi}

Pointwise mutual information (PMI) is a measure of the strength of the association between two words
in a corpus \cite{church89}. Let $p(a)$ and $p(b)$ be the probabilities of observing the words $a$ and
$b$ in a given corpus. Let $p(a, b)$ be the probability of observing $a$ and $b$ together. The PMI
of $a$ and $b$ is defined as follows:

\begin{equation}
\textrm{pmi}(a,b) = \textrm{log} \! \left( \frac{p(a, b)}{p(a)p(b)} \right)
\label{eq:pmi}
\end{equation}

\noindent PMI ranges from negative infinity to positive infinity. Positive values indicate association
and negative values indicate a lack of association. Research suggests that negative PMI values are not useful
in vector space models of semantics \cite{bullinaria07}, so it is common to use positive pointwise mutual
information (PPMI), defined as follows:

\begin{equation}
\textrm{ppmi}(a,b) =
\left\{
\begin{array}{rl}
{\rm pmi}(a,b) & \mbox{if ${\rm pmi}(a,b) > 0$} \\
0 & \mbox{otherwise}
\end{array}
\right.
\label{eq:ppmi}
\end{equation}

\noindent PPMI ranges from zero to positive infinity.

In experiments with our training datasets, we found that accuracy improved when PPMI was forced to
range between zero and one, by applying a logistic function. The logistic function is a sigmoid
function (an {\em S}-shaped range squashing function) with the following equation:

\begin{equation}
f(x) = \frac{1}{1 + e^{-x}}
\end{equation}

\noindent When $x$ is zero, $f(x)$ is 0.5. Positive infinity maps to one and negative infinity maps
to zero. We applied a linear rescaling to the logistic function, as follows:

\begin{equation}
g(x) = \frac{2}{1 + e^{-x}} - 1
\label{eq:gx}
\end{equation}

\noindent The function $g(x)$ maps zero to zero and positive infinity to one. When we apply
$g$ to PPMI and use the natural logarithm (base $e$), the $e^{-x}$ in Equation~\ref{eq:gx}
and the log in Equation~\ref{eq:pmi} cancel out, yielding the following result:

\begin{equation}
g(\textrm{ppmi}(a,b)) =
\left\{
\begin{array}{rl}
\dfrac{2}{1 + \frac{p(a)p(b)}{p(a, b)}} - 1 & \mbox{if $p(a, b) > p(a)p(b)$} \\
0 & \mbox{otherwise}
\end{array}
\right.
\label{eq:gppmi}
\end{equation}

\noindent We use this normalized form of PPMI, $g(\textrm{ppmi}(a,b))$, in our experiments. (Another
approach to normalizing PPMI is given in \citeA{bouma09}.)

The positive pointwise mutual information features are stored in a sparse matrix. The general procedure for
creating a PPMI matrix is described in detail in \citeA{turney10}. In the following experiments, we use the
PPMI matrix from \citeA{turney11}.\footnote{The PPMI matrix is available from the author on request.} It
is a word-context matrix in which the rows correspond to $n$-grams in WordNet and the columns correspond to
unigrams from WordNet, marked {\em left} or {\em right}. There are approximately 114,000 rows in the matrix
and 140,000 columns. The matrix has a density of about 1.2\%.

Let $a$ be an $n$-gram in WordNet, $b$ be a unigram in WordNet, and $h$ be either {\em left} or {\em right}.
Suppose that $a$ corresponds to the $i$-th row in the matrix and $b$, marked with the handedness $h$, corresponds
to the $j$-th column in the matrix. We define $\textrm{PPMI}(a, b, h)$  as the value in the $i$-th row and $j$-th
column of the matrix. This value is the normalized positive pointwise mutual information of observing $b$ on the
$h$ side of $a$ in the Waterloo corpus, where $b$ is either immediately adjacent to $a$ or separated from $a$ by
one or more stop words.\footnote{\citeA{turney11} did not normalize PPMI in their experiments. They used
Equation~\ref{eq:ppmi}, whereas we use Equation~\ref{eq:gppmi} here.} Any word
that is not in WordNet is treated as a stop word. If $a$ does not correspond to a row in the matrix or $b$
(marked $h$) does not correspond to column, then $\textrm{PPMI}(a, b, h)$ is assigned the value zero.

\citeA{turney11} estimated $\textrm{PPMI}(a, b, h)$ by sampling the Waterloo corpus for phrases containing
$a$ and then looking for $b$ on the $h$ side of $a$ in the sampled phrases. Due to this sampling process,
$\textrm{PPMI}(a, b, \textit{left})$ does not necessarily equal $\textrm{PPMI}(b, a, \textit{right})$.
Suppose $a$ is a rare word and $b$ is common. Given $\textrm{PPMI}(a, b, \textit{left})$, when
we sample phrases containing $a$, we are relatively likely to find $b$ in some of these phrases. Given
$\textrm{PPMI}(b, a, \textit{right})$, when we sample phrases containing $b$, we are less likely to find
any phrases containing $a$. Although, in theory, $\textrm{PPMI}(a, b, \textit{left})$ should equal
$\textrm{PPMI}(b, a, \textit{right})$, they are likely to be unequal given a limited sample.

Given a pseudo-unigram, $a\_b$, $\textrm{PPMI}(a\_b, c, h)$ can be found by looking for the
row that corresponds to the bigram $ab$ and the column that corresponds to the unigram $c$, marked $h$.
However, $\textrm{PPMI}(c, a\_b, h)$ is a problem, because the columns of the PPMI matrix
are unigrams, so the bigram $ab$ will never correspond to a column in the matrix. In this case, we
approximate $\textrm{PPMI}(c, a\_b, h)$ by $\textrm{PPMI}(a\_b, c, h')$, where $h'$ is the opposite
handedness of $h$. We never need to compute $\textrm{PPMI}$ when both arguments are pseudo-unigrams,
$\textrm{PPMI}(a\_b, c\_d, h)$.

\subsection{DS: Domain Similarity}
\label{subsec:ds}

Domain similarity (DS) was designed to capture the topic (the field, area, or domain) of a word. The following
experiments use the domain matrix from \citeA{turney12}.\footnote{The domain matrix is available from
the author on request.} To make the domain matrix, \citeA{turney12} first constructed a frequency matrix,
in which the rows correspond to $n$-grams in WordNet and the columns correspond to nouns that were
observed near the row $n$-grams in the Waterloo corpus. The hypothesis was that the nouns near a term
characterize the topics associated with the term. Given an $n$-gram, $a$, the Waterloo corpus was sampled
for phrases containing $a$ and the phrases were processed with a part-of-speech tagger, to identify nouns.
If the noun $b$ was the closest noun to the left or right of $a$, where $a$ corresponds to the \mbox{$i$-th}
row of the frequency matrix and $b$ corresponds to the \mbox{$j$-th} column of the matrix, then the
frequency count for the \mbox{$i$-th} row and \mbox{$j$-th} column was incremented.

The word-context frequency matrix for domain space has about 114,000 rows (WordNet terms) and 50,000
columns (noun contexts, topics), with a density of 2.6\%. The frequency matrix was converted to a PPMI
matrix and then processed with singular value decomposition (SVD).\footnote{The PPMI matrix was
based on Equation~\ref{eq:ppmi}, not Equation~\ref{eq:gppmi}.} The SVD yields three matrices,
$\mathbf{U}$, $\mathbf{\Sigma}$, and $\mathbf{V}$. A term in domain space is represented by a row vector
in $\mathbf{U}_k \mathbf{\Sigma}_k^p$. The parameter $k$ specifies the number of singular values in the
truncated singular value decomposition; that is, $k$ is the number of latent factors in the low-dimensional
representation of the term \cite{landauer97}. We generate $\mathbf{U}_k$ and $\mathbf{\Sigma}_k$
by deleting the columns in $\mathbf{U}$ and $\mathbf{\Sigma}$ corresponding to the smallest
singular values. The parameter $p$ raises the singular values in $\mathbf{\Sigma}_k$ to the power
$p$ \cite{caron01}. When $p$ is zero, all of the $k$ factors have equal weight. When $p$ is one,
each factor is weighted by its corresponding singular value in $\mathbf{\Sigma}$. Decreasing $p$
has the effect of making the similarity measure more discriminating \cite{turney12}.

The similarity of two words in domain space, $\textrm{DS}(a, b, k, p)$, is computed by extracting the
row vectors in $\mathbf{U}_k \mathbf{\Sigma}_k^p$ that correspond to the $n$-grams $a$ and $b$, and then
calculating their cosine. Optimal performance requires tuning the parameters $k$ and $p$ for the task
\cite{bullinaria12,turney12}. For Comp and Decomp, we use the parameter settings given by \citeA{turney12}.
For Super, we generate features with a wide range of parameter settings and let the
supervised learning algorithm decide how to use the features.

If either $a$ or $b$ does not correspond to a row in $\mathbf{U}_k \mathbf{\Sigma}_k^p$, then
$\textrm{DS}(a, b, k, p)$ is set to zero. Since the rows in the domain matrix correspond to $n$-grams
in WordNet, pseudo-unigrams do not present a problem. Given a pseudo-unigram, $a\_b$, we compute
$\textrm{DS}(a\_b, c, k, p)$ by extracting the row vectors in $\mathbf{U}_k \mathbf{\Sigma}_k^p$
that correspond to the bigram $ab$ and the unigram $c$ and then calculating the cosine of the vectors.

\subsection{FS: Function Similarity}
\label{subsec:fs}

Function similarity (FS) was designed to capture the function of a word (its usage, role, or relationship).
The following experiments use the function matrix from \citeA{turney12}.\footnote{The function matrix is
available from the author on request.} It is similar to the domain matrix, except the context is based on
verbal patterns, instead of nearby nouns. The hypothesis was that the functional role of a word is
characterized by the patterns that relate the word to nearby verbs.

The word-context frequency matrix for function space has about 114,000 rows (WordNet terms) and
50,000 columns (verb pattern contexts, functional roles), with a density of 1.2\%. The
frequency matrix was converted to a PPMI matrix and smoothed with SVD.\footnote{The PPMI matrix was
based on Equation~\ref{eq:ppmi}, not Equation~\ref{eq:gppmi}.} The similarity of two words
in function space, $\textrm{FS}(a, b, k, p)$, is computed by extracting the row vectors in
$\mathbf{U}_k \mathbf{\Sigma}_k^p$ that correspond to the $n$-grams $a$ and $b$, and then
calculating their cosine.

\section{Algorithms}
\label{sec:algorithms}

This section presents the algorithms, Comp, Decomp, and Super. Comp and Decomp use fast
unsupervised algorithms to generate initial lists of candidates \cite{turney12}.
Super uses a slower supervised algorithm to refine the initial lists from Comp or Decomp
\cite{turney13}. Super handles both composition and decomposition using the same
algorithm, but it requires different training datasets and it builds different models
for the two different tasks.

For composition, Comp takes a bigram $ab$ as input and generates a ranked list of $max_{\rm u}$
unigrams as output. Then Super takes the bigram $ab$ and the list of $max_{\rm u}$ unigrams as
input and generates a new ranking of the list as output.

For decomposition, Decomp takes a unigram $a$ as input and generates a ranked list of $max_{\rm b}$
bigrams as output. Then Super takes the unigram $a$ and the list of $max_{\rm b}$ bigrams as
input and generates a new ranking of the list as output.

\subsection{Comp}
\label{subsec:comp}

\begin{itemize}[itemsep=1pt,parsep=1pt,topsep=4pt,partopsep=1pt]    

\item Input: a bigram, $ab$.

\item Output: a ranked list of $max_{\rm u}$ unigrams.

\end{itemize}

\noindent Let $ab$ be a bigram and let $c$ be a unigram, such that $a$, $b$, and $c$ are any of the
73,000 WordNet unigrams. Comp uses a scoring function, $\textrm{score}_{\rm u}(ab, c)$,
to estimate the quality of the unigram $c$ considered as a semantic composition of the bigram $ab$. The
score is based on the geometric mean of DS (domain similarity) and FS (function similarity).
The geometric mean is only suitable for positive numbers, but DS and FS can be negative.
We suppress these negative values for Comp and Decomp, but we allow them for Super, since
the supervised learning algorithm should be able to make use of them. Let $\textrm{nn}(x)$
({\em nn} for {\em nonnegative}) be $x$ when $x > 0$ and let it be zero otherwise. We define
$\textrm{score}_{\rm u}(ab, c)$ as follows:

\begin{equation}
\textrm{score}_{\rm u}(ab, c) =
\sqrt{ \textrm{nn}(\textrm{DS}(a, c, k_{\rm d}, p_{\rm d}))
\cdot \textrm{nn}(\textrm{FS}(b, c, k_{\rm f}, p_{\rm f})) }
\label{eq:scoreu}
\end{equation}

\noindent This heuristic score is based on the idea that, when $ab$ is a noun-modifier
bigram, the noun $c$ should be in the same general domain as the modifier, $a$, and have
the same general function as the head noun, $b$. For example, suppose $ab$ is
{\em red salmon} and $c$ is {\em sockeye}. It seems reasonable to say that {\em sockeye}
is in the domain of things that are {\em red} and {\em sockeye} has the same functional
role as {\em salmon}.

For a given bigram, $ab$, Comp calculates $\textrm{score}_{\rm u}(ab, c)$ for every
unigram, $c$, in the 73,000 WordNet unigrams. The candidate unigrams are then sorted
in order of decreasing score and the top $max_{\rm u}$ highest scoring candidates
are the output of Comp.

We use the Perl Data Langague (PDL) package to rapidly calculate $\textrm{score}_{\rm u}(ab, c)$.
The scores can be calculated quickly by processing all of the candidate unigrams together
in a matrix, instead of working with each individual candidate as a single vector.

Table~\ref{tab:comp-parameters} shows the parameter settings we use for Comp in the
following experiments. The values of the first four parameters were copied from
\citeA{turney12}. The last parameter, $max_{\rm u}$, was set to 2000, based on a small number of
trials using the standard composition training dataset. Larger values for $max_{\rm u}$ tend
to yield improved accuracy, after applying Super, but this comes at the cost of increased
execution time for Super.

\begin{table}
\begin{center}
\scalebox{0.9}{
\begin{tabular}{lll}
\hline
Parameter          & Description                                          & Value \\
\hline
$k_{\rm d}$        & number of latent factors for domain space            &  800   \\
$p_{\rm d}$        & exponent for singular values in domain space         &    0.3 \\
$k_{\rm f}$        & number of latent factors for function space          &  100   \\
$p_{\rm f}$        & exponent for singular values in function space       &    0.6 \\
$max_{\rm u}$      & maximum candidate unigrams per target bigram         & 2000   \\
\hline
\end{tabular}
} 
\end{center}
\caption {Parameters for Comp.}
\label{tab:comp-parameters}
\end{table}

\subsection{Decomp}
\label{subsec:decomp}

\begin{itemize}[itemsep=1pt,parsep=1pt,topsep=4pt,partopsep=1pt]    

\item Input: a unigram, $a$.

\item Output: a ranked list of $max_{\rm b}$ bigrams.

\end{itemize}

\noindent Let $a$ be a unigram and let $bc$ be a bigram, such that $a$, $b$, and $c$ are any of
the 73,000 Wordnet unigrams. Decomp uses a scoring function, $\textrm{score}_{\rm m}(a, b)$, to
estimate the quality of the unigram $b$ considered as a modifier in the bigram $bc$, where $bc$
is a semantic decomposition of $a$. We define $\textrm{score}_{\rm m}(a, b)$ as follows:

\begin{equation}
\textrm{score}_{\rm m}(a, b) =
\textrm{nn}(\textrm{DS}(a, b, k_{\rm d}, p_{\rm d}))
\cdot (\textrm{PPMI}(a, b, \mathit{left}) + \textrm{PPMI}(a, b, \mathit{right}))
\label{eq:scorem}
\end{equation}

\noindent Decomp also uses a scoring function, $\textrm{score}_{\rm h}(a, c)$, to estimate
the quality of the unigram $c$ considered as a head noun in the bigram $bc$. We define
$\textrm{score}_{\rm h}(a, c)$ as follows:

\begin{equation}
\textrm{score}_{\rm h}(a, c) =
\textrm{nn}(\textrm{FS}(a, c, k_{\rm f}, p_{\rm f}))
\cdot (\textrm{PPMI}(a, c, \mathit{left}) + \textrm{PPMI}(a, c, \mathit{right}))
\label{eq:scoreh}
\end{equation}

\noindent Like Comp, Decomp uses DS (domain similarity) for
modifiers and FS (function similarity) for heads. The PPMI terms in Equations
\ref{eq:scorem} and \ref{eq:scoreh} are designed to give more weight to candidate
modifiers and heads that are observed in the corpus near the target unigram, $a$.
For example, suppose $a$ is {\em sockeye}, $b$ is {\em red}, and $c$ is {\em salmon}.
In a large corpus, if we sample phrases that contain {\em sockeye}, we would expect
a few of these phrases to contain either {\em red} ({\em the sockeye was a deep red})
or {\em salmon} ({\em their favourite salmon was sockeye}). The scoring functions,
$\textrm{score}_{\rm m}(a, b)$ and $\textrm{score}_{\rm h}(a, c)$, have no preference
about whether $b$ and $c$ appear to the {\em right} or {\em left} of $a$. If they
appear on both sides of $a$, that is better than appearing on only one side.

Decomp scores every unigram, $b$, in the 73,000 Wordnet unigrams with
$\textrm{score}_{\rm m}(a, b)$. The unigrams are then sorted in order of decreasing
score and the top $max_{\rm m}$ highest scoring unigrams are considered as
candidate modifiers. Every unigram, $c$, is scored with $\textrm{score}_{\rm h}(a, c)$
and the top $max_{\rm h}$ highest scoring unigrams are taken as candidate heads.
Decomp then generates $max_{\rm m} \cdot max_{\rm h}$ bigrams, $bc$, by combining
each candidate modifier, $b$, with each candidate head, $c$. Finally, these
$max_{\rm m} \cdot max_{\rm h}$ bigrams are scored with $\textrm{score}_{\rm b}(a, bc)$,
defined as follows:

\begin{equation}
\textrm{score}_{\rm b}(a, bc) =
\textrm{score}_{\rm m}(a, b) \cdot \textrm{score}_{\rm h}(a, c)
\cdot \textrm{LBF}(b, c) \cdot \textrm{PPMI}(b,c,\mathit{right})
\label{eq:scoreb}
\end{equation}

\noindent The scored bigrams are sorted in order of decreasing score and the top
$max_{\rm b}$ highest scoring candidates are the output of Decomp. The LBF term
in Equation~\ref{eq:scoreb} is designed to give more weight to candidate
bigrams that occur more frequently. The PPMI term increases the score when
$c$ is often found to the {\em right} of $b$ in the corpus.

Table~\ref{tab:decomp-parameters} shows the parameter settings we use for Decomp in the
following experiments. The first four parameters are set the same as for Comp in
Table~\ref{tab:comp-parameters}. The remaining three, $max_{\rm m}$, $max_{\rm h}$, and
$max_{\rm b}$, were set based on a small number of trials using the standard decomposition
training dataset. Larger values tend to improve accuracy at the cost of increased
execution time for Super.

\begin{table}
\begin{center}
\scalebox{0.9}{
\begin{tabular}{lll}
\hline
Parameter          & Description                                          & Value \\
\hline
$k_{\rm d}$        & number of latent factors for domain space            &  800   \\
$p_{\rm d}$        & exponent for singular values in domain space         &    0.3 \\
$k_{\rm f}$        & number of latent factors for function space          &  100   \\
$p_{\rm f}$        & exponent for singular values in function space       &    0.6 \\
$max_{\rm m}$      & maximum candidate modifiers per target unigram       & 1000   \\
$max_{\rm h}$      & maximum candidate heads per target unigram           & 1000   \\
$max_{\rm b}$      & maximum candidate bigrams per target unigram         & 2000   \\
\hline
\end{tabular}
} 
\end{center}
\caption {Parameters for Decomp.}
\label{tab:decomp-parameters}
\end{table}

The equations for Decomp are more complex than the equation for Comp, because Decomp
is exploring a larger set of candidates. Given a target bigram, $ab$, Comp considers
73,000 possible unigram compositions. Given a target unigram, $a$, Decomp considers
5,300,000,000 (73,000 squared) candidate bigram decompositions. The two algorithms
share the core functions, DS and FS, but Decomp requires the additional functions,
PPMI and LBF, in order to handle the larger set.

With $max_{\rm m}$ and $max_{\rm h}$ set to 1000, $max_{\rm m} \cdot max_{\rm h}$ is
1,000,000. Thus $\textrm{score}_{\rm b}(a, bc)$ is applied to one million candidate
bigrams for each input target unigram, $a$. This may seem to be a large set of candidates,
but it is considerably smaller than the 5,300,000,000 possible bigrams, and it is
also smaller than the 314,843,401 bigrams in the Google Web 1T 5-gram (Web1T5) dataset
\cite{brants06}. This is why Decomp splits the task into two steps: First select subsets
of the candidate modifiers and heads independently, then evaluate bigrams formed from
these relatively small subsets.

\subsection{Super}
\label{subsec:super}

\begin{itemize}[itemsep=1pt,parsep=1pt,topsep=4pt,partopsep=1pt]    

\item Input: a list of triples of the form $\langle a, b, c \rangle$.

\item Output: a ranking of the input list.

\end{itemize}

\noindent Super uses a supervised learning algorithm to refine the lists that it gets from Comp
or Decomp. Comp takes a target bigram, $ab$, as input and generates $max_{\rm u}$ unigrams,
$c_1, c_2, \ldots$ as output. Super views this as $max_{\rm u}$ triples of the form
$\langle a, b, c_i \rangle$, where $i$ ranges from one to $max_{\rm u}$. Decomp takes a target
unigram, $a$, as input and generates $max_{\rm b}$ bigrams, $b_1 c_1, b_2 c_2, \ldots$ as output.
Super views this as $max_{\rm b}$ triples of the form $\langle a, b_i, c_i \rangle$, where $i$
ranges from one to $max_{\rm b}$. In both cases, the task of Super is to learn to
rank triples. Super uses the same feature vectors to represent these triples, regardless
of whether the triples come from Comp or Decomp.

The first step for Super is to represent each triple with a feature vector. Let $\langle a, b, c \rangle$
be a triple, where $a$, $b$, and $c$ are unigrams. Super generates three LUF features, one for each
unigram, $a$, $b$, and $c$. There are six LBF features, one for each possible pair of unigrams,
$\langle a, b \rangle$, $\langle a, c \rangle$, $\langle b, a \rangle$, $\langle b, c \rangle$,
$\langle c, a \rangle$, and $\langle c, b \rangle$. For PPMI, there are six possible pairs of unigrams
and two possible values of $h$ ({\em left} and {\em right}), hence there are twelve features. For DS
and FS, the order of the pairs does not matter, because cosine is symmetric
(${\rm cos}(\mathbf{a},\mathbf{b}) = {\rm cos}(\mathbf{b},\mathbf{a})$), so there are only three pairs
to consider, $\langle a, b \rangle$, $\langle a, c \rangle$, and $\langle b, c \rangle$. However, we
explore ten values of $k$ and eleven values of $p$, so there are $3 \times 10 \times 11 = 330$ features.
This gives a total of 681 features, as summarized in Table~\ref{tab:super-features}. These are the same
features as in \citeA{turney13}, except for the six new LBF features.

\begin{table}
\begin{center}
\scalebox{0.9}{
\begin{tabular}{llr}
\hline
Feature type                  & Description                           & Instances \\
\hline
$\textrm{LUF}(a)$             & log unigram frequency                 & 3    \\
$\textrm{LBF}(a, b)$          & log bigram frequency                  & 6    \\
$\textrm{PPMI}(a, b, h)$      & positive pointwise mutual information & 12   \\
$\textrm{DS}(a, b, k, p)$     & domain similarity                     & 330  \\
$\textrm{FS}(a, b, k, p)$     & function similarity                   & 330  \\
\hline
total                         &                                       & 681  \\
\hline
\end{tabular}
} 
\end{center}
\caption {Features for Super.}
\label{tab:super-features}
\end{table}

Since Super is supervised, it requires a training dataset. For example, the standard composition
training dataset contains 154 target bigrams (see Table~\ref{tab:datasets}). Given that $max_{\rm u}$
is 2000, after running Comp on this training dataset, we have $154 \times 2000 = 308,\!000$ triples,
and thus $308,\!000$ training feature vectors.

We do not use all of these feature vectors to train the model. First, we drop any target bigrams
for which there are no solutions. For example, we see in Table~\ref{tab:datasets-samples} that the target
bigram {\em foot lever} has the solutions {\em pedal} and {\em treadle}. If neither of these unigrams
appear among the 2000 candidates in the output of Comp, then we remove all of the 2000 triples containing
{\em foot lever} from the training dataset. For the standard composition training dataset, this step
leaves us with 135 target bigrams, for a total of $135 \times 2000 = 270,\!000$ training triples.

Second, we adjust the class ratio. Table~\ref{tab:datasets} shows that there are 1.5 possible
solutions per target bigram in the standard composition training dataset, but not all of these solutions
appear in the 2000 candidates per target that are generated by Comp. For the 135 target bigrams
that survive the first filtering step above, there are 1.33 solutions per target bigram that
actually occur in the 2000 candidates per target from Comp. This gives us $135 \times 1.33 = 180$
triples that we label as class 1 (correct solutions). The remaining $270,\!000 - 180 = 269,\!820$
triples are labeled as class 0 (incorrect candidates). This class imbalance makes learning difficult
\cite{japkowicz02}. Therefore, for each of the 135 target bigrams, we select all triples from class 1
and we randomly sample triples from class 0, until we have $ratio_{01}$ triples in class 0 for
every triple in class 1. In the following experiments, $ratio_{01}$ is 30, so we have 180 triples
in class 1 and $180 \times 30 = 5,\!400$ triples in class 0, for a total of $5,\!400 + 180 = 5,\!580$ triples.

The above numbers use the standard composition training dataset as an example, but the same
basic method is applied to all four datasets: First, drop targets that lack solutions. Second,
adjust the class ratio. These steps only apply when training. No modifications are made to the
testing datasets. 

Super uses the sequential minimal optimization (SMO) support vector machine (SVM) as implemented in Weka
\cite{platt98,witten11}.\footnote{Weka is available at \url{http://www.cs.waikato.ac.nz/ml/weka/}.}
The kernel is a normalized third-order polynomial. Weka provides probability estimates for
the classes by fitting the outputs of the SVM with logistic regression models. After training, we apply
Super to the testing dataset. For each testing target, we rank the 2000 candidates in descending order of
the probability that they belong in class 1, as estimated by the SVM.

Table~\ref{tab:super-parameters} shows the parameter settings that we use for Super in the
following experiments. The Super parameter values are copied from \citeA{turney13}, except for
$ratio_{01}$, which was set based on a small number of trials using the standard composition training
dataset. The same parameter settings are used for all four datasets.

\begin{table}
\begin{center}
\scalebox{0.9}{
\begin{tabular}{lll}
\hline
Parameter          & Description                                          & Value \\
\hline
$K$                & range of values for $k$ in DS and FS ($|K| = 10$)    & $\{100, 200, 300, \ldots, 1000\}$ \\
$P$                & range of values for $p$ in DS and FS ($|P| = 11$)    & $\{0.0, 0.1, 0.2, \ldots, 1.0\}$  \\
$deg$              & degree of polynomial kernel in SVM                   & 3  \\
$ratio_{01}$       & ratio of class 0 to class 1 for SVM                  & 30 \\
\hline
\end{tabular}
} 
\end{center}
\caption {Parameters for Super.}
\label{tab:super-parameters}
\end{table}

Supervised learning is able to construct more accurate scoring functions than Equations \ref{eq:scoreu}
to \ref{eq:scoreb}, but Super requires training data and it executes more slowly. Super uses 681 features,
whereas Comp uses only two features and Decomp uses only eight features. (The features in Comp and Decomp
are included among the 681 features of Super. See Equations \ref{eq:scoreu} to \ref{eq:scoreb}.) Most of
the computation time of Super is spent calculating the extra features. By combining Comp and Decomp with
Super, we are able to get the speed advantage of the unsupervised heuristics and the accuracy advantage
of supervised learning.

\section{Experiments with Composition}
\label{sec:composition}

This section describes the experiments with Comp and Super on the standard and holistic
composition datasets.

\subsection{Evaluation with the Standard Composition Dataset}
\label{subsec:comp-standard}

Table~\ref{tab:red-salmon} uses the target bigram {\em red salmon} to illustrate the process of
Comp and Super. (This is the first target in the standard composition testing dataset; see
Table~\ref{tab:datasets-samples}.) The solution unigram, {\em sockeye}, is marked with an asterisk.
Among the top 2000 candidates in the output of Comp, {\em sockeye} is ranked 445th, with a score
of 0.167, according to Equation~\ref{eq:scoreu}. When the 2000 candidates from Comp are rescored by
Super, {\em sockeye} rises up to rank third, with a score of 0.5, according to the SVM model.

\begin{table}
\begin{center}
\scalebox{0.9}{
\begin{tabular}{lrlcclc}
\hline
             &          & \multicolumn{2}{c}{Comp}       & & \multicolumn{2}{c}{Super} \\
                          \cline{3-4}                        \cline{6-7}
             & Rank     & Term       & Score             & & Term       & Score    \\
\hline
Target:      &          & red salmon &                   & & red salmon &          \\
\hline
Candidates:  &       1  & red        & 0.504             & & striped    & 0.573    \\
             &       2  & teal       & 0.374             & & herring    & 0.569    \\
             &       3  & green      & 0.366             & & sockeye *  & 0.500    \\
             &       4  & plum       & 0.329             & & hooknose   & 0.459    \\
             &       5  & blue       & 0.325             & & rockfish   & 0.451    \\
             &       6  & cerulean   & 0.313             & & red        & 0.341    \\
             & $\cdots$ &            & $\cdots$          & &            & $\cdots$ \\
             &     445  & sockeye *  & 0.167             & & croaker    & 0.011    \\
             & $\cdots$ &            & $\cdots$          & &            & $\cdots$ \\
             &    2000  & eros       & 0.117             & & eros       & 0.000    \\
\hline
\end{tabular}
} 
\end{center}
\caption {Candidate compositions of {\em red salmon}, ranked by Comp and Super.}
\label{tab:red-salmon}
\end{table}

Table~\ref{tab:eval-comp-standard} summarizes the performance of Comp and Super on the
standard composition testing dataset, containing 351 target bigrams. The row labeled
{\em mean rank in 2000 candidates} is the mean rank of first correct answer, if any, in
the 2000 guesses. The row labeled {\em median rank in 2000 candidates} is the median rank
of first correct answer, if any, in the 2000 guesses. The rows labeled {\em percent in top N}
give the percentage of the targets for which a correct solution is one of the top $N$
guesses. If there are two or more solutions in the top $N$ guesses, we use the rank of
the first solution. The row {\em percent in 2000 candidates} is the percentage of the target
bigrams for which there is a correct answer anywhere in the $max_{\rm u}$ (2000) candidates
generated by Comp. The final row, {\em number of candidates considered}, shows the
number of possibilities considered by Comp and Super. 

\begin{table}
\begin{center}
\scalebox{0.9}{
\begin{tabular}{lSS}  
\hline
Evaluation metric &  \multicolumn{1}{c}{Comp}          &  \multicolumn{1}{c}{Super} \\
\hline
Mean rank in 2000 candidates               &   198     &    60     \\
Median rank in 2000 candidates             &    26     &     5     \\
Percent in top 1                           &     6.6   &    17.7   \\
Percent in top 10                          &    30.2   &    50.4   \\
Percent in top 100                         &    56.7   &    77.8   \\
Percent in 2000 candidates                 &    84.0   &    84.0   \\
\hline
Number of candidates considered & \multicolumn{1}{c}{72,924}  & \multicolumn{1}{c}{2000} \\
\hline
\end{tabular}
} 
\end{center}
\caption {Performance of Comp and Super on the standard composition dataset.}
\label{tab:eval-comp-standard}
\end{table}

When calculating the mean and median rank, we only include the targets for which a correct answer is
among the 2000 guesses, since the rank is not defined for Super when the correct answer is not
among the 2000 guesses. Thus the mean and median only cover the 84.0\% of the 351 targets.
This makes the mean and median slightly misleading, because an algorithm could improve its mean and
median ranks by refusing to guess unless it is very confident. The other four evaluation metrics cover
100\% of the 351 targets. We include all six metrics in order to provide a broad perspective on the
performance of the algorithms. Our preferred evaluation metric is {\em percent in top 100},
the percentage of targets for which a correct answer is one of top 100 guesses. By this metric,
Comp and Super, working together, achieve a score of 77.8\%.

\subsection{Comparison with Baselines}
\label{subsec:comp-baselines}

We compare Comp and Super with three baselines, vector addition, element-wise multiplication,
and the holistic approach. Let $ab$ be a bigram and let $c$ be a unigram. Let $\mathbf{a}$, $\mathbf{b}$,
and $\mathbf{c}$ be the corresponding context vectors. With vector addition, we score the triple
$\langle a, b, c \rangle$ using ${\rm cos}(\mathbf{a} + \mathbf{b}, \mathbf{c})$, the cosine of the angle
between $\mathbf{a} + \mathbf{b}$ and $\mathbf{c}$ \cite{landauer97}. With element-wise
multiplication, the triple is scored by ${\rm cos}(\mathbf{a} \odot \mathbf{b}, \mathbf{c})$
\cite{mitchell08,mitchell10}. With the holistic approach, we treat $ab$ as a pseudo-unigram. Let
$\mathbf{d}$ be the context vector for the pseudo-unigram $a\_b$. We score the triple using
${\rm cos}(\mathbf{d}, \mathbf{c})$.

For vector addition, we used the domain matrix (see Section~\ref{subsec:ds}), since this
matrix had the best performance with addition in \citeA{turney12}. We used the training dataset to
optimize the parameters $k$ and $p$ of the smoothed matrix, $\mathbf{U}_k \mathbf{\Sigma}_k^p$.
The best results were obtained with $k$ set to 1000 and $p$ set to $-0.1$.

For element-wise multiplication, we used the PPMI matrix (see Section~\ref{subsec:ppmi}),
since this matrix has no negative elements. \citeA{turney12} pointed out that element-wise
multiplication is not suitable when the vectors contain negative elements. The DS and FS
matrices contain negative elements, due to the truncated singular value decomposition.
\citeA{turney12} suggested a modified form of element-wise multiplication, to address this
issue, but we found that it did not scale up to the number of vectors we have in our testing
dataset (351 targets $\times$ 2000 candidates per target = 702,000 testing dataset vectors).
With the PPMI matrix, there are no parameters to tune.

For the holistic approach, we used the mono matrix from \citeA{turney12}, since this matrix had
the best performance with the holistic approach in \citeA{turney12}. The mono matrix was formed by
merging the domain and function matrices. See \citeA{turney12} for details. We used the training
dataset to optimize the parameters $k$ and $p$ of the smoothed matrix, $\mathbf{U}_k \mathbf{\Sigma}_k^p$.
The best results were obtained with $k$ set to 1300 and $p$ set to $-0.5$.

Since \citeA{turney12} showed that the geometric mean of domain similarity and function
similarity performed better than vector addition and element-wise multiplication on the
noun-modifier composition recognition task, we decided to apply addition and multiplication
to the output of Comp, instead of applying them to the full set of 73,000 unigrams. The intention
was to give vector addition and element-wise multiplication the benefit of preprocessing
by Comp. On the other hand, \citeA{turney12} found that the holistic approach was
more accurate than all other approaches; therefore we applied the holistic approach to the
whole set of 73,000 unigrams, with no preproessing by Comp.

Table~\ref{tab:eval-comp-baselines} shows the results. When we look at {\em percent in top 1} and
{\em percent in top 10} in Table~\ref{tab:eval-comp-baselines}, addition and multiplication seem
to impair the ranking done by Comp. For {\em percent in top 100}, there seems to be some benefit
to addition and multiplication. However, Super (77.8\%) is significantly better than addition (66.1\%)
and multiplication (60.4\%), according to Fisher's Exact Test at the 95\% confidence level. The
difference between Super (77.8\%) and the holistic approach (78.9\%) is not signficant. For
{\em percent in top 10}, the difference between Super (50.4\%) and the holistic approach (54.4\%)
is also not signficant. For {\em percent in top 1}, the holistic approach (25.1\%) is significantly
better than Super (17.7\%).

\begin{table}
\begin{center}
\scalebox{0.9}{
\begin{tabular}{lSSSSS}  
\hline
Evaluation metric & \multicolumn{1}{c}{Comp} & \multicolumn{1}{c}{Super}
& \multicolumn{1}{c}{Addition} & \multicolumn{1}{c}{Multiplication} & \multicolumn{1}{c}{Holistic} \\
\hline
Mean rank in 2000 candidates               &   198     &    60     &  103    &   165    &   81   \\
Median rank in 2000 candidates             &    26     &     5     &   22    &    21    &    6   \\
Percent in top 1                           &     6.6   &    17.7   &    0.3  &     0.0  &   25.1 \\
Percent in top 10                          &    30.2   &    50.4   &   29.9  &    27.6  &   54.4 \\
Percent in top 100                         &    56.7   &    77.8   &   66.1  &    60.4  &   78.9 \\
Percent in 2000 candidates                 &    84.0   &    84.0   &   84.0  &    84.0  &   92.9 \\
\hline
Number of candidates considered & \multicolumn{1}{c}{72,924} & \multicolumn{1}{c}{2000}
& \multicolumn{1}{c}{2000} & \multicolumn{1}{c}{2000} & \multicolumn{1}{c}{72,924} \\
\hline
\end{tabular}
} 
\end{center}
\caption {Performance of Comp and Super compared to baselines.}
\label{tab:eval-comp-baselines}
\end{table}

Note that the holistic approach can only handle target bigrams included in the 36,000 bigrams
in WordNet, whereas Super can handle any target bigram that contains unigrams from the 73,000 unigrams
in WordNet (that is, 5,300,000,000 bigrams). Because we chose to base our vocabulary on WordNet,
the target bigrams in the standard composition dataset are restricted to bigrams that appear in WordNet.
Since our datasets (Section~\ref{sec:datasets}) and our features (Section~\ref{sec:features}) are based
on the vocabulary of WordNet, the holistic approach seems more useful than it would appear if the testing
dataset were not derived from WordNet.

The holistic approach cannot not scale up to 5,300,000,000 bigrams. Comp and Super scale up and attain
accuracy near the level of the holistic approach, although the testing dataset favours the holistic
approach.

\subsection{Evaluation with the Holistic Composition Dataset}
\label{subsec:comp-holistic}

Given the standard composition dataset, Super learns from 5,580 triples (see Section~\ref{subsec:super}),
derived from 154 WordNet synonym sets (see Table~\ref{tab:datasets}). Super learns
from the expert knowledge that is embedded in WordNet. We would like to be able to train Super
without using this kind of expertise.

Past work with adjective-noun bigrams has shown that we can use holistic bigram vectors to train a
supervised regression model \cite{guevara10,baroni10}. \citeA{turney13} adapted this approach
for supervised classification applied to composition recognition. The purpose of the two holistic
datasets (see Section~\ref{sec:datasets}) is to investigate whether we can apply holistic training to
composition and decomposition generation. The holistic datasets were generated without using
WordNet synomym sets. To construct the holistic datasets, we used WordNet as a source of bigrams, but
we ignored the rich information that WordNet provides about these bigrams, such as their synonyms,
hypernyms, hyponyms, meronyms, and glosses.

Table~\ref{tab:eval-comp-holistic} shows the performance of Comp and Super on the holistic
compostion dataset. For ease of comparison, the table also shows the performance on the
standard composition dataset. The training and testing subsets of the holistic composition dataset
have the same sizes as the corresponding training and testing subsets of the standard
composition dataset (154 training targets and 351 testing targets). However, we see that Comp and
Super achieve much better results on the holistic dataset than on the standard dataset. This
suggests that the holistic dataset is not nearly as challenging as the dataset based on WordNet
synonym sets.

\begin{table}
\begin{center}
\scalebox{0.9}{
\begin{tabular}{lSScSS}  
\hline
\hspace{\fill}Training and testing: & \multicolumn{2}{c}{Holistic} & & \multicolumn{2}{c}{Standard} \\
\cline{2-3} \cline{5-6}
Evaluation metric &  \multicolumn{1}{c}{Comp}         &  \multicolumn{1}{c}{Super} &
&  \multicolumn{1}{c}{Comp}         &  \multicolumn{1}{c}{Super} \\
\hline
Mean rank in 2000 candidates               &    85     &     6    & &   198     &    60     \\
Median rank in 2000 candidates             &     5     &     2    & &    26     &     5     \\
Percent in top 1                           &    10.5   &    26.8  & &     6.6   &    17.7   \\
Percent in top 10                          &    59.8   &    86.6  & &    30.2   &    50.4   \\
Percent in top 100                         &    80.3   &    93.7  & &    56.7   &    77.8   \\
Percent in 2000 candidates                 &    94.6   &    94.6  & &    84.0   &    84.0   \\
\hline
Number of candidates considered & \multicolumn{1}{c}{114,101} & \multicolumn{1}{c}{2000} &
& \multicolumn{1}{c}{72,924} & \multicolumn{1}{c}{2000} \\
\hline
\end{tabular}
} 
\end{center}
\caption {Comp and Super on the holistic and standard composition datasets.}
\label{tab:eval-comp-holistic}
\end{table}

Table~\ref{tab:eval-comp-mixed-holistic} looks at the performance with various combinations
of testing and training datasets. Although Comp and Super work together, the table only
shows the results for Super, in order to make the table easier to read. The two columns that
are labeled {\em both} use \citeS{daume07} domain adaptation algorithm for training. This algorithm
allows us to train Super with both the standard training dataset and the holistic training dataset.
\citeS{daume07} algorithm is a general strategy for merging datasets that have somewhat different
statistical distributions of class labels, such as our standard and holistic datasets.

\begin{table}
\begin{center}
\scalebox{0.9}{
\begin{tabular}{lSSScSSS}  
\hline
\hspace{\fill}Testing: & \multicolumn{3}{c}{Holistic} & & \multicolumn{3}{c}{Standard} \\
\cline{2-4} \cline{6-8}
Evaluation metric \hspace{\fill} Training: & \multicolumn{1}{c}{Holistic} & \multicolumn{1}{c}{Standard} &
\multicolumn{1}{c}{Both} & & \multicolumn{1}{c}{Standard} & \multicolumn{1}{c}{Holistic} & \multicolumn{1}{c}{Both} \\
\hline
Mean rank in 2000 candidates               &     6     &   14    &   6    & &    60     &   130    &   56     \\
Median rank in 2000 candidates             &     2     &    2    &   2    & &     5     &    17    &    5     \\
Percent in top 1                           &    26.8   &   29.6  &  26.2  & &    17.7   &     7.1  &   17.1   \\
Percent in top 10                          &    86.6   &   76.1  &  87.2  & &    50.4   &    34.8  &   51.9   \\
Percent in top 100                         &    93.7   &   92.3  &  93.7  & &    77.8   &    64.4  &   76.6   \\
Percent in 2000 candidates                 &    94.6   &   94.6  &  94.6  & &    84.0   &    84.0  &   84.0   \\
\hline
Number of candidates considered & \multicolumn{1}{c}{2000} & \multicolumn{1}{c}{2000}
& \multicolumn{1}{c}{2000} & & \multicolumn{1}{c}{2000} & \multicolumn{1}{c}{2000} & \multicolumn{1}{c}{2000} \\
\hline
\end{tabular}
} 
\end{center}
\caption {Performance of Super with various combinations of training and testing datasets.}
\label{tab:eval-comp-mixed-holistic}
\end{table}

Table~\ref{tab:eval-comp-mixed-holistic} shows that the model learned from standard training
(with WordNet synonym sets) carries over well to holistic testing. Focusing on {\em percent in top 100},
standard training achieves 92.3\% on holistic testing, which is not significantly different
from the 93.7\% obtained with holistic training (Fisher's Exact Test at the 95\% confidence level).
On the other hand, the model learned from holistic training (without WordNet synonym sets) does not
carry over well to standard training. Holistic training only achieves 64.4\% on standard testing,
whereas standard training achieves 77.8\%, which is significantly higher.

Table~\ref{tab:eval-comp-mixed-holistic} suggests that there is not much benefit to merging
the holistic and standard training datasets with \citeS{daume07} domain adaptation algorithm.
Training on the standard training dataset alone achieves results that are as good as merging
the datasets.

\subsection{Evaluation with the Adjective-Noun Dataset}
\label{subsec:comp-adj-noun}

Here we experiment with the 620 adjective-noun phrases of \citeA{dinu13} (see Table~\ref{tab:dinu13}
in Section~\ref{subsec:rel-comp}).\footnote{Georgiana Dinu, Nghia The Pham, and Marco Baroni gave us a
copy of their 620 adjective-noun phrases, the corresponding noun solutions for each noun phrase,
and their vocabulary of 21,000 nouns.} Table~\ref{tab:eval-comp-adj-noun} shows the performance of
Comp and Super on the adjective-noun dataset. For training, we merged the standard training and testing
composition datasets ($154 + 351 = 505$ targets) and removed any target bigrams that appeared in the
adjective-noun dataset ($505 - 115 = 390$ targets). For testing, we used all 620 adjective-noun phrases.

\begin{table}
\begin{center}
\scalebox{0.9}{
\begin{tabular}{lSS}  
\hline
Evaluation metric &  \multicolumn{1}{c}{Comp}          &  \multicolumn{1}{c}{Super} \\
\hline
Mean rank in 2000 candidates               &   232     &    117     \\
Median rank in 2000 candidates             &    29     &     10     \\
Percent in top 1                           &     7.7   &     11.1   \\
Percent in top 10                          &    28.7   &     40.6   \\
Percent in top 100                         &    50.6   &     62.6   \\
Percent in 2000 candidates                 &    78.9   &     78.9   \\
\hline
Number of candidates considered & \multicolumn{1}{c}{72,924}  & \multicolumn{1}{c}{2000} \\
\hline
\end{tabular}
} 
\end{center}
\caption {Performance of Comp and Super on the adjective-noun dataset.}
\label{tab:eval-comp-adj-noun}
\end{table}

There is a slight decrease in the performance with the adjective-noun phrases
(Table~\ref{tab:eval-comp-adj-noun}), compared to the noun-modifier dataset
(Table~\ref{tab:eval-comp-standard}). We hypothesize that this is because we deliberately sought
highly compositional phrases for the noun-modifier dataset, using WordNet glosses as heuristic
clues (see Section~\ref{sec:datasets}). It seems that the phrases in the adjective-noun dataset
are less compositional than the phrases in the noun-modifier dataset.

The 620 adjective-noun phrases were introduced by \citeA{dinu13}, but they were used more recently,
and with better results, by \citeA{li14}. Table~\ref{tab:li14} is a copy of their results
(this is Table~3 in their paper). They evaluated seven different models on the adjective-noun
dataset. The {\em reduction} column indicates the algorithm used to reduce the dimensionality
of the matrix, nonnegative matrix factorization (NMF) or singular value decomposition (SVD).
The {\em dim} column indicates the number of factors in the dimensionality reductions. The
{\em rank} column gives the median rank of the solution in the ranked list of candidates.

\begin{table}
\begin{center}
\scalebox{0.9}{
\begin{tabular}{llrr}
\hline
Model    & Reduction   &  Dim   &  Rank \\
\hline
Add      & NMF         &  300   &  113.0 \\
Dil      & NMF         &  300   &  354.5 \\
Mult     & NMF         &  300   &  146.5 \\
Fulladd  & SVD         &  300   &  123.0 \\
Lexfunc  & SVD         &  150   &  117.5 \\
Fulllex  & SVD         &   50   &  394.0 \\
EnetLex  & SVD         &  300   &  108.5 \\
\hline
\end{tabular}
} 
\end{center}
\caption {Performance of models on the adjective-noun dataset, from \citeA{li14}.}
\label{tab:li14}
\end{table}

In Table~\ref{tab:li14}, {\em Add} is vector addition with weights \cite{mitchell08}.
{\em Dil} is the dilation model introduced by \citeA{mitchell08}. {\em Mult} is element-wise
multiplication \cite{mitchell08} with powers as weights \cite{dinu13}. {\em Fulladd} multiplies each
vector by a weight matrix and then adds the resulting weighted vectors \cite{guevara10}. {\em Lexfunc}
represents adjectives by matrices and nouns by vectors \cite{baroni10}; the adjective matrix is a kind of
{\em lexical function} that modifies the noun vector. {\em Fulllex} represents every component unigram,
whether noun or adjective, with both a vector and a matrix \cite{socher12}. {\em Enetlex} is
a lexical function model, like {\em Lexfunc}, where the model is trained using
elastic-net regression \cite{li14}. 

All of the models in Table~\ref{tab:li14} have weights or parameters that are tuned
or learned on training data. The training takes a holistic approach. Given a bigram, $ab$,
the models are trained to predict the vector of the pseudo-unigram, $a\_b$. 

In order to compare our results on this dataset with the results of \citeA{li14},
we need to make two adjustments to our experimental setup. First, we consider a vocabulary
of 73,000 unigrams, but \citeA{li14} consider 21,000 nouns. For a fair comparison,
we restrict Comp to the same 21,000 nouns. FilterComp is a modified version of Comp
that filters the output of Comp to remove any unigrams that are not among the 21,000
nouns of \citeA{li14}. Second, we only calculate the mean and median rank for the
targets that have solutions among the top 2000 candidates, because Super does not
rank the other candidates. Since \citeA{li14} use a single pass approach, they
rank all of the candidates. To make our median comparable to their median, we
give a rank to all candidates. If a target does not have a solution among the top
2000 candidates, we assume that the rank of the solution is the worst possible rank;
that is, the rank of the last candidate, 21,098.

Table~\ref{tab:eval-filtercomp-adj-noun} shows the performance of FilterComp
and Super with these two adjustments. Filtering makes the task easier by removing
irrelevant candidates, which tends to improve the results, but our adjustment to
the mean and median calculation has a negative impact on these evaluation metrics.
Fortunately the median is a robust statistic that is only slightly affected
by the adjustment. The mean, on the other hand, is greatly changed, for the worse.

\begin{table}
\begin{center}
\scalebox{0.9}{
\begin{tabular}{lSS}  
\hline
Evaluation metric &  \multicolumn{1}{c}{FilterComp}    &  \multicolumn{1}{c}{Super} \\
\hline
Mean rank in 21,098 candidates             &  3611     &  3515     \\
Median rank in 21,098 candidates           &    52     &    13     \\
Percent in top 1                           &    10.0   &    14.0   \\
Percent in top 10                          &    32.4   &    47.7   \\
Percent in top 100                         &    57.1   &    70.6   \\
Percent in 2000 candidates                 &    83.7   &    83.7   \\
\hline
Number of candidates considered & \multicolumn{1}{c}{21,098}  & \multicolumn{1}{c}{2000} \\
\hline
\end{tabular}
} 
\end{center}
\caption {Performance of FilterComp and Super on the adjective-noun dataset.}
\label{tab:eval-filtercomp-adj-noun}
\end{table}

With medians of 52 and 13, FilterComp and Super are performing much better than
the best model in Table~\ref{tab:li14}, EnetLex, with a median rank of 108.5.
Recall the discussion in Section~\ref{subsec:rel-comp} about context composition
and similarity composition. All of the models in Table~\ref{tab:li14} involve
context composition, whereas FilterComp and Super use similarity composition.
We believe that this is the main reason for the better performance of FilterComp and Super.

Super is attempting to learn a function, $f$, that maps feature vectors to scalar probabilities.
The majority of the features are various kinds of similarities (see Table~\ref{tab:super-features}).
The models in Table~\ref{tab:li14} are attempting to learn a function, $\mathbf{f}$, that maps to
vectors. The models are trained with pseudo-unigram context vectors. We believe that
pseudo-unigram context vectors are only an approximation, a surrogate for what we
really want to learn. We really want to learn that {\em red salmon} and {\em sockeye}
are synonymous, but instead we learn that {\em red salmon} and {\em red\_salmon} are synonymous.
The results in Section~\ref{subsec:comp-holistic} suggest that {\em red\_salmon}
is a reasonable substitute for {\em sockeye}, but it is not quite as good as the real thing.

Futhermore, Lexfunc and EnetLex must learn a separate model (a unique matrix) for
each adjective in the vocabulary. There are 411 different adjectives in the 620
adjective-noun phrases; thus Lexfunc and EnetLex require at least 411 different
trained models in order to find solution nouns for the 620 targets. FilterComp
and Super use one model for all noun-modifier expressions, both adjective-noun
and noun-noun. They can readily handle adjectives that never appeared in their
training data, unlike Lexfunc and EnetLex.

\section{Experiments with Decomposition}
\label{sec:decomposition}

This section describes the experiments with Decomp and Super on the standard and holistic
decomposition datasets.

\subsection{Evaluation with the Standard Decomposition Dataset}
\label{subsec:decomp-standard}

Decomp first builds a list of $max_{\rm m}$ modifiers using $\textrm{score}_{\rm m}$ and a list
of $max_{\rm h}$ heads using $\textrm{score}_{\rm h}$. Table~\ref{tab:sockeye-mods-heads}
illustrates this first step, using the target unigram {\em sockeye} as an example.
Table~\ref{tab:datasets-samples} shows that {\em sockeye} has several possible solution bigrams:
{\em blueback salmon}, {\em red salmon}, {\em sockeye salmon}, and {\em Oncorhynchus nerka}.
Looking ahead, the solution that will eventually be found is {\em sockeye salmon}. In
Table~\ref{tab:sockeye-mods-heads}, {\em sockeye} is ranked first in the list of candidate modifiers
and {\em salmon} is ranked sixth in the list of candidate heads. (They are marked with asterisks in
the table.)

\begin{table}
\begin{center}
\scalebox{0.9}{
\begin{tabular}{lrlcclc}
\hline
             &          & \multicolumn{2}{c}{Modifiers}  & & \multicolumn{2}{c}{Heads} \\
                          \cline{3-4}                        \cline{6-7}
             & Rank     & Term              & Score      & & Term                   & Score    \\
\hline
Target:      &          & sockeye           &            & & sockeye                &          \\
\hline
Candidates:  &       1  & sockeye *         & 1.996      & & sockeye                & 1.996    \\
             &       2  & coho              & 1.803      & & coho                   & 1.873    \\
             &       3  & chinook           & 1.582      & & chinook                & 1.701    \\
             &       4  & escapement        & 1.492      & & chum                   & 1.508    \\
             &       5  & anadromous        & 1.450      & & trout                  & 1.383    \\
             &       6  & salmon            & 1.440      & & salmon *               & 1.382    \\
             & $\cdots$ &                   & $\cdots$   & &                        & $\cdots$ \\
             &    1000  & gear              & 0.000      & & ward                   & 0.026    \\
\hline
\end{tabular}
} 
\end{center}
\caption {Candidate modifiers and heads for decomposing {\em sockeye}.}
\label{tab:sockeye-mods-heads}
\end{table}

Next, Decomp builds a list of $max_{\rm b}$ bigrams using $\textrm{score}_{\rm b}$ to rank the
combination of each of the $max_{\rm m}$ modifiers with each of the $max_{\rm h}$ heads.
Super then rescores the list of bigrams. Table~\ref{tab:sockeye} continues the {\em sockeye}
example. Decomp ranks {\em sockeye salmon} first. Unfortunately, Super moves the rank
of {\em sockeye salmon} down to the twenty-first choice. (Asterisks mark {\em sockeye salmon}
in the table.)

\begin{table}
\begin{center}
\scalebox{0.9}{
\begin{tabular}{lrlcclc}
\hline
             &          & \multicolumn{2}{c}{Decomp}     & & \multicolumn{2}{c}{Super} \\
                          \cline{3-4}                        \cline{6-7}
             & Rank     & Term              & Score      & & Term                   & Score    \\
\hline
Target:      &          & sockeye           &            & & sockeye                &          \\
\hline
Candidates:  &       1  & sockeye salmon *  & 31.3       & & silver salmon          & 0.970    \\
             &       2  & coho salmon       & 29.1       & & pink salmon            & 0.963    \\
             &       3  & sockeye coho      & 28.2       & & red herring            & 0.941    \\
             &       4  & chinook salmon    & 26.5       & & alaska salmon          & 0.936    \\
             &       5  & chum salmon       & 20.7       & & summer chinook         & 0.786    \\
             &       6  & coho chum         & 20.4       & & commercial salmon      & 0.774    \\
             & $\cdots$ &                   & $\cdots$   & &                        & $\cdots$ \\
             &      21  & sockeye fry       & 14.2       & & sockeye salmon *       & 0.374    \\
             & $\cdots$ &                   & $\cdots$   & &                        & $\cdots$ \\
             &    2000  & seabird foraging  & 0.5        & & zooplankton predation  & 0.000    \\
\hline
\end{tabular}
} 
\end{center}
\caption {Candidate decompositions of {\em sockeye}, ranked by Decomp and Super.}
\label{tab:sockeye}
\end{table}

The second guess of Super in Table~\ref{tab:sockeye} is {\em pink salmon}, which seems like
a good decomposition for {\em sockeye}, but it is not a member of the WordNet synonym set for {\em sockeye}.
{\em Pink salmon} ({\em Oncorhynchus gorbuscha}) and {\em red salmon} ({\em Oncorhynchus nerka, sockeye})
are distinct species. {\em Coho salmon} ({\em silver salmon}, {\em Oncorhynchus kisutch}),
{\em chinook salmon} ({\em Oncorhynchus tshawytscha}), and {\em chum salmon} ({\em Oncorhynchus keta})
are other species of salmon.

Table~\ref{tab:eval-decomp-standard} summarizes the performance of Decomp and Super
on the standard decomposition testing dataset, containing 355 target unigrams.
According to our preferred evaluation metric, {\em percent in top 100}, Decomp and Super,
working together, achieve a score of 50.7\% on this dataset. This is below the 77.8\% that Comp
and Super reach on the standard composition testing dataset, but Comp and Super only consider
73,000 candidates, whereas Decomp and Super consider $5.3 \times 10^9$. Decomposition
is a more difficult task than composition.

\begin{table}
\begin{center}
\scalebox{0.9}{
\begin{tabular}{lSS}  
\hline
Evaluation metric &  \multicolumn{1}{c}{Decomp}       &  \multicolumn{1}{c}{Super} \\
\hline
Mean rank in 2000 candidates               &   250     &    80     \\
Median rank in 2000 candidates             &    82     &    24     \\
Percent in top 1                           &     4.2   &     5.6   \\
Percent in top 10                          &    14.9   &    22.0   \\
Percent in top 100                         &    34.9   &    50.7   \\
Percent in 2000 candidates                 &    63.7   &    63.7   \\
\hline
Number of candidates considered & \multicolumn{1}{c}{$\rule{0pt}{11pt} 5.3 \times 10^9$} 
& \multicolumn{1}{c}{2000} \\   
\hline
\end{tabular}
} 
\end{center}
\caption {Performance of Decomp and Super on the standard decomposition dataset.}
\label{tab:eval-decomp-standard}
\end{table}

\subsection{Comparison with Baselines}
\label{subsec:decomp-baselines}

As with composition (Section~\ref{subsec:comp-baselines}), we compare Decomp and Super with
three baselines, vector addition, element-wise multiplication, and the holistic approach.
Addition and multiplication take the output of Decomp and rescore it. The holistic approach
scores all bigrams in WordNet, treating the bigrams as pseudo-unigrams.
Table~\ref{tab:eval-decomp-baselines} shows the results. Addition and multiplication perform
significantly worse than Decomp and Super according to {\em percent in top 1}, {\em top 10},
and {\em top 100} (Fisher's Exact Test, 95\% confidence level). The holistic approach performs
significantly better than Decomp and Super according to {\em percent in top 1}, {\em top 10},
and {\em top 100} (Fisher's Exact Test, 95\% confidence level).

\begin{table}
\begin{center}
\scalebox{0.9}{
\begin{tabular}{lSSSSS}  
\hline
Evaluation metric &  \multicolumn{1}{c}{Decomp}       &  \multicolumn{1}{c}{Super}
& \multicolumn{1}{c}{Addition} & \multicolumn{1}{c}{Multiplication} & \multicolumn{1}{c}{Holistic} \\
\hline
Mean rank in 2000 candidates               &   250     &    80     &   433     &   543     &    58    \\
Median rank in 2000 candidates             &    82     &    24     &   238     &   303     &     3    \\
Percent in top 1                           &     4.2   &     5.6   &     0.8   &     0.6   &    34.4  \\
Percent in top 10                          &    14.9   &    22.0   &     6.5   &     6.8   &    63.4  \\
Percent in top 100                         &    34.9   &    50.7   &    22.3   &    19.4   &    83.1  \\
Percent in 2000 candidates                 &    63.7   &    63.7   &    63.7   &    63.7   &    94.6  \\
\hline
Number of candidates considered & \multicolumn{1}{c}{$\rule{0pt}{11pt} 5.3 \times 10^9$}
& \multicolumn{1}{c}{2000} & \multicolumn{1}{c}{2000} & \multicolumn{1}{c}{2000}
& \multicolumn{1}{c}{36,149} \\   
\hline
\end{tabular}
} 
\end{center}
\caption {Performance of Decomp and Super compared to baselines.}
\label{tab:eval-decomp-baselines}
\end{table}

Note that the holistic approach can only decompose a unigram into a bigram if the bigram is one
of the 36,000 bigrams in WordNet, whereas Decomp and Super can decompose a unigram into any bigram
that contains unigrams from the 73,000 unigrams in WordNet (5,300,000,000 bigrams). Although
the holistic approach performs better in Table~\ref{tab:eval-decomp-baselines}, Decomp and Super
are considering a much larger set of candidates.

For a more fair comparison, we can restrict Decomp to the same 36,000 bigrams as the holistic
approach. FilterDecomp is a modified version of Decomp that simply filters the output of Decomp
to remove any bigrams that are not in WordNet. Table~\ref{tab:eval-decomp-restrict} compares
FilterDecomp and Super to the holistic approach. Now FilterDecomp and Super approach the performance
of the holistic approach. The difference in {\em percent in top 1}, 23.9\% for Super versus
34.4\% for the holistic approach, is statistically significant, but the differences in {\em top 10}
(58.3\% versus 63.4\%) and {\em top 100} (78.6\% versus 83.1\%) are not significant (Fisher's Exact
Test, 95\% confidence level). Also, the performance of Super on the decomposition dataset
(78.6\% in Table~\ref{tab:eval-decomp-restrict}) is now near its performance on the composition
dataset (77.8\% in Table~\ref{tab:eval-comp-standard}). This suggests that the main difference
between the composition and decomposition tasks is simply the number of candidates that must be
considered.

\begin{table}
\begin{center}
\scalebox{0.9}{
\begin{tabular}{lSSS}  
\hline
Evaluation metric &  \multicolumn{1}{c}{FilterDecomp}       &  \multicolumn{1}{c}{Super}
& \multicolumn{1}{c}{Holistic} \\
\hline
Mean rank in 2000 candidates               &   36    &   20    &    58    \\
Median rank in 2000 candidates             &    9    &    4    &     3    \\
Percent in top 1                           &   18.9  &   23.9  &    34.4  \\
Percent in top 10                          &   42.8  &   58.3  &    63.4  \\
Percent in top 100                         &   75.2  &   78.6  &    83.1  \\
Percent in 2000 candidates                 &   82.8  &   82.8  &    94.6  \\
\hline
Number of candidates considered & \multicolumn{1}{c}{36,149} & \multicolumn{1}{c}{2000}
& \multicolumn{1}{c}{36,149} \\
\hline
\end{tabular}
} 
\end{center}
\caption {Performance of FilterDecomp and Super compared with Holistic.}
\label{tab:eval-decomp-restrict}
\end{table}

The point of Table~\ref{tab:eval-decomp-restrict} is only to show that Decomp and Super
are competitive with the performance of the holistic approach, contrary to appearances
in Table~\ref{tab:eval-decomp-baselines}. We would not actually want to use FilterDecomp
in a realistic application. One problem is that 36,000 bigrams are not enough bigrams to
serve as miniature definitions for 73,000 unigrams.

\subsection{Evaluation with the Holistic Decomposition Dataset}
\label{subsec:decomp-holistic}

Table~\ref{tab:eval-decomp-holistic} shows the performance of Decomp and Super on the holistic
decomposition dataset. The previous results for the standard decomposition dataset
(Table~\ref{tab:eval-decomp-standard}) are copied here to make comparison easier. As with the
composition task (see Table~\ref{tab:eval-comp-holistic}), the results show that the holistic
dataset is not nearly as challenging as the dataset based on WordNet synonym sets.

\begin{table}
\begin{center}
\scalebox{0.9}{
\begin{tabular}{lSScSS}  
\hline
\hspace{\fill}Training and testing: & \multicolumn{2}{c}{Holistic} & & \multicolumn{2}{c}{Standard} \\
\cline{2-3} \cline{5-6}
Evaluation metric &  \multicolumn{1}{c}{Decomp}         &  \multicolumn{1}{c}{Super} &
&  \multicolumn{1}{c}{Decomp}       &  \multicolumn{1}{c}{Super} \\
\hline
Mean rank in 2000 candidates               &    198     &    10    & &   250     &    80     \\
Median rank in 2000 candidates             &     39     &     2    & &    82     &    24     \\
Percent in top 1                           &     11.0   &    33.0  & &     4.2   &     5.6   \\
Percent in top 10                          &     28.2   &    69.0  & &    14.9   &    22.0   \\
Percent in top 100                         &     54.1   &    81.7  & &    34.9   &    50.7   \\
Percent in 2000 candidates                 &     83.1   &    83.1  & &    63.7   &    63.7   \\
\hline
Number of candidates considered & \multicolumn{1}{c}{$\rule{0pt}{11pt} 5.3 \times 10^9$}
& \multicolumn{1}{c}{2000} & & \multicolumn{1}{c}{$\rule{0pt}{11pt} 5.3 \times 10^9$}
& \multicolumn{1}{c}{2000} \\  
\hline
\end{tabular}
} 
\end{center}
\caption {Decomp and Super on the holistic and standard decomposition datasets.}
\label{tab:eval-decomp-holistic}
\end{table}

Table~\ref{tab:eval-decomp-mixed-holistic} gives results for various combinations of testing
and training datasets. Here we see some benefit to merging the two training datasets,
using \citeS{daume07} domain adaptation algorithm (see the columns labeled {\em both}).
However, if we focus on {\em percent in top 100}, the differences between training with the
standard dataset and the merged dataset (80.3\% versus 82.3\% with holistic testing and
50.7\% versus 52.7\% with standard testing) are not significant. Training on the standard
training dataset alone achieves results that are as good as merging the datasets. We still
do not see a significant benefit from using the holistic dataset.

\begin{table}
\begin{center}
\scalebox{0.9}{
\begin{tabular}{lSSScSSS}  
\hline
\hspace{\fill}Testing: & \multicolumn{3}{c}{Holistic} & & \multicolumn{3}{c}{Standard} \\
\cline{2-4} \cline{6-8}
Evaluation metric \hspace{\fill} Training: & \multicolumn{1}{c}{Holistic} & \multicolumn{1}{c}{Standard} &
\multicolumn{1}{c}{Both} & & \multicolumn{1}{c}{Standard} & \multicolumn{1}{c}{Holistic} & \multicolumn{1}{c}{Both} \\
\hline
Mean rank in 2000 candidates               &  10    & 16    &  7    & &  80    & 158    & 68    \\
Median rank in 2000 candidates             &   2    &  3    &  2    & &  24    &  70    & 20    \\
Percent in top 1                           &  33.0  & 22.5  & 41.4  & &   5.6  &   1.7  &  5.9  \\
Percent in top 10                          &  69.0  & 61.7  & 72.7  & &  22.0  &  10.1  & 22.5  \\
Percent in top 100                         &  81.7  & 80.3  & 82.3  & &  50.7  &  38.6  & 52.7  \\
Percent in 2000 candidates                 &  83.1  & 83.1  & 83.1  & &  63.7  &  63.7  & 63.7  \\
\hline
Number of candidates considered & \multicolumn{1}{c}{2000} & \multicolumn{1}{c}{2000}
& \multicolumn{1}{c}{2000} & & \multicolumn{1}{c}{2000} & \multicolumn{1}{c}{2000} & \multicolumn{1}{c}{2000} \\
\hline
\end{tabular}
} 
\end{center}
\caption {Performance of Super with various combinations of training and testing datasets.}
\label{tab:eval-decomp-mixed-holistic}
\end{table}

\section{Discussion and Analysis}
\label{sec:discussion}

In Section~\ref{subsec:comp-standard}, we evaluted Comp and Super by comparing their output to
WordNet synonym sets. For the bigrams in the standard compositon dataset, the top 100 most highly
ranked unigrams included a WordNet synonym 77.8\% of the time (Table~\ref{tab:eval-comp-standard}).
In Section~\ref{subsec:decomp-standard}, we evaluated Decomp and Super with WordNet. For the
standard decomposition dataset, the top 100 bigrams included a WordNet synonym 50.7\% of the time
(Table~\ref{tab:eval-decomp-standard}). Together, Comp and Super explore 73,000 candidate
unigram compositions for a bigram. Decomp and Super explore 5,300,000,000 candidate bigram
decompositions for a unigram. Given that Decomp and Super explore a much larger space than
Comp and Super, it is encouraging that Decomp and Super are able to achieve 50.7\%.

In Sections \ref{subsec:comp-baselines} and \ref{subsec:decomp-baselines}, we compared
Comp, Decomp, and Super to three baselines, vector addition, element-wise multiplication,
and the holistic approach. Given the output of Comp, addition and multiplication are able to
improve the percentage of solutions found in the top 100 candidates, but Super is able to
achieve a significantly larger improvement (Table~\ref{tab:eval-comp-baselines}). Given the
output of Decomp, addition and multiplication impair the ranking of Decomp, whereas
Super improves the ranking of Decomp (Table~\ref{tab:eval-decomp-baselines}). However,
the holistic approach seems to achieve better results than Comp, Decomp, and Super.

Comp and Super, working together, achieve 50.4\% in the top 10 and 77.8\% in the top 100,
whereas the holistic approach achieves 54.4\% and 78.9\% (Table~\ref{tab:eval-comp-baselines}),
but the differences in these scores are not statistically significant. Decomp and Super
at first seem to be performing substantially below the holistic approach (Table~\ref{tab:eval-decomp-baselines}),
but this does not take into account that Decomp and Super explore 5,300,000,000 candidate bigram
decompositions, whereas the holistic approach, due to its scaling problems, can only explore
36,000 decompositions. Once we correct for this (Table~\ref{tab:eval-decomp-restrict}),
the differences in the top 10 (58.3\% for FilterDecomp and Super versus 63.4\% for the holistic
approach) and the top 100 (78.6\% versus 83.1\%) are not statistically significant.

Given the good performance of the holistic baseline, it is natural to consider using holistic
datasets as a relatively inexpensive way to train a supervised system \cite{guevara10,baroni10,turney13}.
The strength of Super is that it uses a compositional approach, so it can scale up, but
its weakness is that it requires training. The strength of the holistic approach is that
it works well without training, but its weakness is that it does not scale up. The hope
is that training Super with holistic datasets will give us both of the strengths: a compositional
approach with relatively inexpensive holistic training.

We explored holistic training in Sections \ref{subsec:comp-holistic} and \ref{subsec:decomp-holistic}.
In the experiments where the training and testing datasets are the same (both are standard
or both are holistic), we observed that the holistic datasets are much easier
to master than the standard WordNet-based datasets (Tables \ref{tab:eval-comp-holistic}
and \ref{tab:eval-decomp-holistic}). Understanding the experiments where the training and testing
datasets are different (Tables \ref{tab:eval-comp-mixed-holistic} and \ref{tab:eval-decomp-mixed-holistic})
takes more effort. Here we present a model that may give some insight into the {\em cross-domain}
results (that is, the results when training on one kind of domain, standard or holistic,
and testing on another). We are particularly interested in training on a holistic dataset
(because it is relatively inexpensive) and then testing on a standard dataset (because
it is more representative of potential real-world applications).

Suppose that Super is given a target $t$ and generates a list of the top $N$ candidate
compositions or decompositions for $t$. Let $P_{\rm hh}(t)$ be the probability that there
is a solution for $t$ among the $N$ candidates generated by Super, where the subscript hh
on $P_{\rm hh}(t)$ indicates holistic training and holistic testing. Let $P_{\rm hs}(t)$ be the
probability that there is a solution for the target $t$ among the top $N$ candidates of Super, where
the subscript hs indicates holistic training and standard testing. Let $Q_{\rm ss}(t)$
be the probability that there is a solution for the target $t$ among the top $N$ candidates
of the holistic approach, where ss indicates standard training and standard testing.
The holistic approach does not require training in the usual sense, but we do use
the standard training set to tune the parameters, $k$ and $p$. We model $P_{\rm hs}(t)$ as
$P_{\rm hh}(t) \cdot Q_{\rm ss}(t)$. That is, the probability that Super can transfer what was
learned in the holistic domain to targets sampled from the standard domain, $P_{\rm hs}(t)$, can
be estimated by multiplying the probability that Super can successfully emulate the holistic approach,
$P_{\rm hh}(t)$, with the probability that the holistic approach can successfully handle the
standard domain, $Q_{\rm ss}(t)$.

We can view the results with holistic training and holistic testing (Tables \ref{tab:eval-comp-holistic}
and \ref{tab:eval-decomp-holistic}) as providing us with estimates of the probability that Super
can successfully emulate the holistic approach, $P_{\rm hh}(t)$. Likewise, we can view the results with the
holistic approach, given standard training and testing (Tables \ref{tab:eval-comp-baselines} and
\ref{tab:eval-decomp-baselines}), as providing us with estimates of the probability that the holistic
approach can find solutions to the WordNet-based targets, $Q_{\rm ss}(t)$. If Super is trained on
the holistic datasets and then evaluated on the standard datasets, then we can estimate its
probability of success, $P_{\rm hs}(t)$, by $P_{\rm hh}(t) \cdot Q_{\rm ss}(t)$.

Table~\ref{tab:model-cross-composition} compares the cross-domain model with the actual
observed performance of Comp and Super on the composition datasets:

\begin{enumerate}[itemsep=1pt,parsep=1pt,topsep=4pt,partopsep=1pt,label*=\Alph*.]    

\item The values in column A show the performance of Super when trained and tested using the
holistic composition dataset. We interpret the percentages in this column as estimated probabilities
for $P_{\rm hh}(t)$ with $N = \{1, 10, 100, 2000\}$.

\item This column gives the performance of the holistic approach when trained and tested using
the standard composition dataset. We interpret the percentages in this column as estimated probabilities
for $Q_{\rm ss}(t)$ with varying values of $N$.

\item The numbers in column C are the product of the corresponding values in columns A and B,
based on interpreting the percentages as probabilities. For example, for $N = 1$, we have
$0.268 \times 0.251 = 0.067$. We interpret the percentages in this column as the predictions
of our model, $P_{\rm hh}(t) \cdot Q_{\rm ss}(t)$, with varying values of $N$.

\item This column shows the performance of Super when trained on the holistic composition
dataset and tested on the standard composition dataset. This column gives the observed values
of $P_{\rm hs}(t)$, which we may compare to the calculated values in column C.

\end{enumerate}

\noindent The predictions of the model (column C) do not exactly match the observed
values (column D), but the model seems reasonable as a first-order approximation.

\begin{table}
\begin{center}
\scalebox{0.9}{
\begin{tabular}{lSSSS}  
\hline
Composition & \multicolumn{1}{c}{A} & \multicolumn{1}{c}{B} & \multicolumn{1}{c}{C} & \multicolumn{1}{c}{D} \\
Evaluation metric & \multicolumn{1}{c}{Super} & \multicolumn{1}{c}{Holistic}
& \multicolumn{1}{c}{Super} & \multicolumn{1}{c}{Super} \\
\hline
Percent in top 1            & 26.8  & 25.1  &  6.7  &  7.1  \\
Percent in top 10           & 86.6  & 54.4  & 47.1  & 34.8  \\
Percent in top 100          & 93.7  & 78.9  & 73.9  & 64.4  \\
Percent in 2000 candidates  & 94.6  & 92.9  & 87.9  & 84.0  \\
\hline
Training dataset & \multicolumn{1}{c}{holistic} & \multicolumn{1}{c}{standard}
& \multicolumn{1}{c}{holistic} & \multicolumn{1}{c}{holistic} \\
Testing dataset & \multicolumn{1}{c}{holistic} & \multicolumn{1}{c}{standard}
& \multicolumn{1}{c}{standard} & \multicolumn{1}{c}{standard} \\
Source of percentages & \multicolumn{1}{c}{Table~\ref{tab:eval-comp-holistic}}
& \multicolumn{1}{c}{Table~\ref{tab:eval-comp-baselines}}
& \multicolumn{1}{c}{A $\times$ B}
& \multicolumn{1}{c}{Table~\ref{tab:eval-comp-mixed-holistic}} \\
Observed or calculated? & \multicolumn{1}{c}{observed} & \multicolumn{1}{c}{observed}
& \multicolumn{1}{c}{calculated} & \multicolumn{1}{c}{observed} \\
\hline
\end{tabular}
} 
\end{center}
\caption {A model of the cross-domain performance of Super with the composition datasets.}
\label{tab:model-cross-composition}
\end{table}

We cannot apply this model immediately to the performance of Decomp and Super on the
decomposition datasets, because the performance of the holistic approach in Table~\ref{tab:eval-decomp-baselines}
is based on partial search. The holistic approach only considers 36,000 possible candidates,
but we need to base $Q_{\rm ss}(t)$ on full search, considering $5.3 \times 10^9$ candidates.
This is a problem, because the holistic approach will not scale up to this size. We would
need to build a term-context matrix with $5.3 \times 10^9$ rows, which is considerably
beyond the current state of the art. Therefore we approximate the expected performance of
the holistic approach with full search by extrapolating from our results with partial and full search
for Decomp and Super.

Table~\ref{tab:model-holistic-decomposition} gives our estimated probabilities for
the holistic approach, $Q_{\rm ss}(t)$, if it were able to scale up to full search:

\begin{enumerate}[itemsep=1pt,parsep=1pt,topsep=4pt,partopsep=1pt,label*=\Alph*.]    

\item The values in column A show the observed performance of Decomp and Super
with full search ($5.3 \times 10^9$ candidates), using the standard training and
testing decomposition datasets.

\item The values in this column show the observed performance of Decomp and Super
with partial search (36,000 candidates), using the standard training and
testing decomposition datasets.

\item This column gives the ratio of column A to column B. Our assumption is that
we should expect to have the same ratio for the holistic approach.

\item The values in this column are the obverved values for the holistic approach
with partial search (36,000 candidates), using the standard training and
testing decomposition datasets.

\item The values in the final column are the expected values for the holistic approach
with full search ($5.3 \times 10^9$ candidates), using the standard training and
testing decomposition datasets.

\end{enumerate}

\noindent We cannot actually observe the values in column E, given the scaling problems
of the holistic approach, but they seem to be reasonable estimates.

\begin{table}
\begin{center}
\scalebox{0.9}{
\begin{tabular}{lSSSSS}  
\hline
Decomposition & \multicolumn{1}{c}{A} & \multicolumn{1}{c}{B} & \multicolumn{1}{c}{C}
& \multicolumn{1}{c}{D} & \multicolumn{1}{c}{E} \\
Evaluation metric & \multicolumn{1}{c}{Super} & \multicolumn{1}{c}{Super}
& \multicolumn{1}{c}{Super} & \multicolumn{1}{c}{Holistic}
& \multicolumn{1}{c}{Holistic} \\
\hline
Percent in top 1            &  5.6   & 23.9   & 0.234   & 34.4   &  8.1  \\
Percent in top 10           & 22.0   & 58.3   & 0.377   & 63.4   & 23.9  \\
Percent in top 100          & 50.7   & 78.6   & 0.645   & 83.1   & 53.6  \\
Percent in 2000 candidates  & 63.7   & 82.8   & 0.769   & 94.6   & 72.8  \\
\hline
Search space for candidates & \multicolumn{1}{c}{full} & \multicolumn{1}{c}{partial}
& \multicolumn{1}{c}{ratio} & \multicolumn{1}{c}{partial} & \multicolumn{1}{c}{full} \\
Source of numbers & \multicolumn{1}{c}{Table~\ref{tab:eval-decomp-standard}}
& \multicolumn{1}{c}{Table~\ref{tab:eval-decomp-restrict}}
& \multicolumn{1}{c}{A $/$ B} & \multicolumn{1}{c}{Table~\ref{tab:eval-decomp-restrict}}
& \multicolumn{1}{c}{C $\times$ D} \\
Observed or calculated? & \multicolumn{1}{c}{observed} & \multicolumn{1}{c}{observed}
& \multicolumn{1}{c}{calculated} & \multicolumn{1}{c}{observed} & \multicolumn{1}{c}{calculated} \\
\hline
\end{tabular}
} 
\end{center}
\caption {A model of holistic performance on the decomposition dataset with full search.}
\label{tab:model-holistic-decomposition}
\end{table}

Now that we have estimates for $Q_{\rm ss}(t)$ with full search, we can apply the cross-domain model
to the performance of Decomp and Super on the decomposition datasets. Table~\ref{tab:model-cross-decomposition}
gives the results:

\begin{enumerate}[itemsep=1pt,parsep=1pt,topsep=4pt,partopsep=1pt,label*=\Alph*.]    

\item The values show the performance of Super when trained and tested using the
holistic decomposition dataset. We interpret the percentages as estimates
for $P_{\rm hh}(t)$.

\item This column gives the expected performance of the holistic approach, if it could extended
to full search. We interpret the percentages as estimates for $Q_{\rm ss}(t)$.

\item The numbers in column C are the product of the corresponding values in columns A and B.
We interpret the percentages in this column as the predictions of our model, $P_{\rm hh}(t) \cdot Q_{\rm ss}(t)$.

\item This column shows the performance of Super when trained on the holistic decomposition
dataset and tested on the standard decomposition dataset. This column gives the observed values
of $P_{\rm hs}(t)$, which we may compare to the calculated values in column C.

\end{enumerate}

\noindent The predictions of the model (column C) do not exactly match the observed
values (column D), but the model seems to be a reasonable approximation.

\begin{table}
\begin{center}
\scalebox{0.9}{
\begin{tabular}{lSSSS}  
\hline
Decomposition & \multicolumn{1}{c}{A} & \multicolumn{1}{c}{B} & \multicolumn{1}{c}{C} & \multicolumn{1}{c}{D} \\
Evaluation metric & \multicolumn{1}{c}{Super} & \multicolumn{1}{c}{Holistic}
& \multicolumn{1}{c}{Super} & \multicolumn{1}{c}{Super} \\
\hline
Percent in top 1                & 33.0   &  8.1   &  2.7   &  1.7  \\
Percent in top 10               & 69.0   & 23.9   & 16.5   & 10.1  \\
Percent in top 100              & 81.7   & 53.6   & 43.8   & 38.6  \\
Percent in 2000 candidates      & 83.1   & 72.8   & 60.5   & 63.7  \\
\hline
Training dataset & \multicolumn{1}{c}{holistic} & \multicolumn{1}{c}{standard}
& \multicolumn{1}{c}{holistic} & \multicolumn{1}{c}{holistic} \\
Testing dataset & \multicolumn{1}{c}{holistic} & \multicolumn{1}{c}{standard}
& \multicolumn{1}{c}{standard} & \multicolumn{1}{c}{standard} \\
Source of percentages & \multicolumn{1}{c}{Table~\ref{tab:eval-decomp-holistic}}
& \multicolumn{1}{c}{Table~\ref{tab:model-holistic-decomposition}}
& \multicolumn{1}{c}{A $\times$ B}
& \multicolumn{1}{c}{Table~\ref{tab:eval-decomp-mixed-holistic}} \\
Observed or calculated? & \multicolumn{1}{c}{observed} & \multicolumn{1}{c}{calculated}
& \multicolumn{1}{c}{calculated} & \multicolumn{1}{c}{observed} \\
\hline
\end{tabular}
} 
\end{center}
\caption {A model of cross-domain performance of Super with the decomposition datasets.}
\label{tab:model-cross-decomposition}
\end{table}

The holistic approach is unsupervised. Although training data is useful for parameter
tuning, a larger training dataset is unlikely to have much impact on $Q_{\rm ss}(t)$.
More training data may improve $P_{\rm hh}(t)$, but the best we can hope for is $P_{\rm hh}(t) = 1$.
Thus our model predicts that $P_{\rm hs}(t)$ cannot be greater than $Q_{\rm ss}(t)$,
and we are not likely to see big improvements in $Q_{\rm ss}(t)$. On the other hand,
training Super with WordNet, $P_{\rm ss}(t)$, is already achieving performance
near $Q_{\rm ss}(t)$. However, our model does not take into account what might
be possible with domain adaptation \cite{daume07}. 

In summary, we can gain some insight into the cross-domain results in Sections
\ref{subsec:comp-holistic} and \ref{subsec:decomp-holistic} (Tables \ref{tab:eval-comp-mixed-holistic}
and \ref{tab:eval-decomp-mixed-holistic}) by viewing the observed performance of
Super, $P_{\rm hs}(t)$, as being the product of two things: the ability of Super
to emulate the holistic approach, $P_{\rm hh}(t)$, and the ability of the holistic
approach to handle targets and solutions based on WordNet, $Q_{\rm ss}(t)$.

\section{Limitations and Future Work}
\label{sec:future}

The main limitation of the work described here is that we have focused our attention
on noun-modifier bigrams and noun unigrams. A natural next step would be to consider
subject-verb bigrams, verb-object bigrams, or subject-verb-object trigrams.

We plan to extend this approach to recognizing when one sentence is a paraphrase of another.
To achieve this goal, we propose to adapt the {\em dynamic pooling} approach introduced by
\citeA{socher11}. Eventually, we hope to be able to move beyond sentence paraphrase recognition
to sentence paraphrase generation. 

\citeA{turney13} has demonstrated that the features we have used here (Section~\ref{sec:features})
work well for recognizing semantic relations; for example, we can recognize that
the semantic relation between {\em mason} and {\em stone} is analogous to the
semantic relation between {\em carpenter} and {\em wood}. Analogy generation is
another possibility for further work.

We believe that improved performance can be obtained by adding more features.
Features based on third-order tensors might be useful for recognizing and generating
paraphrases \cite{baroni2010b}. With the right features, it should be possible
to surpass the accuracy of the holistic approach.

Comp and Decomp rely on hand-crafted score functions (Section~\ref{sec:algorithms}).
The advantage of these hand-crafted functions is that they use only eight features,
which makes them much faster than using the 681 features of Super. One way to
avoid hand-crafting would be to use aggressive feature selection with Super.
Instead of using Comp and Decomp for the first pass through the candidates, we
could use a highly pruned version of Super. We expect that the pruning will reduce
accuracy, so there will still be a benefit from a second pass of Super using the
full feature set. 

Another area for future work is more experimentation with domain adaptation and
holistic training. Although our model (Section~\ref{sec:discussion}) suggests that
there are limits to the value of holistic training, it may be possible to overcome
these limits with domain adaptation \cite{daume07}. 

\section{Conclusions}
\label{sec:conclusions}

The main weakness of the recognition task is that the degree of challenge in the task
depends on the given list of candidates. A critic can always claim that the given list
was too easy; the distractors were too unlike the solution. The generation task avoids
this criticism. Furthermore, the ability to generate solutions opens up opportunities for new
applications, beyond the applications that are possible with the ability to recognize
solutions.

Following the example of \citeA{dinu13} and \citeA{li14}, we have extended distributional
models of composition from the task of recognition to the task of generation. 
Our results on their adjective-noun dataset suggest that similarity composition
is a better approach to generating compositions than context composition
(Section~\ref{subsec:comp-adj-noun}). The experiments with holistic training
support the hypothesis that the models of \citeA{dinu13} and \citeA{li14} are
limited by their reliance on holistic pseudo-unigram training (Section~\ref{subsec:comp-holistic}).

We have also extended distributional models to the task of decomposition generation.
Our results indicate that decomposition is considerably more difficult than
composition, due to the larger search space, which increases the difficulty of
finding the solution among the candidates (Section~\ref{subsec:decomp-baselines}).

By the stringent criterion that allows only one guess ({\em percent
in top 1}), we achieved an accuracy of 17.7\% on composition generation
(Table~\ref{tab:eval-comp-standard}) and 5.6\% on decomposition generation
(Table~\ref{tab:eval-decomp-standard}). By the more relaxed criterion that allows
100 guesses, the accuracy on composition is 77.8\% and the accuracy on decomposition
is 50.7\%. We do not know what accuracy an average human would have on these tasks, but
generating a solution that agrees with WordNet is difficult. Many of the words in WordNet
are unfamiliar and also the developers of WordNet are likely to have missed many reasonable
possibilities for the WordNet synonym sets.

We introduced a simple cross-domain model of holistic training and WordNet-based testing
(Section~\ref{sec:discussion}), which gives some insight into the limitations of holistic
training. With standard training, our models are achieving performance near the level
of the holistic approach (Tables \ref{tab:eval-comp-baselines} and \ref{tab:eval-decomp-restrict}).
With new features (beyond the features in Section~\ref{sec:features}), we expect that Super
will be able to surpass the performance of the holistic approach. (Super already surpasses
the holistic approach in that Super can scale up in a way that is not possible for any
non-compositional approach.) However, the simple cross-domain model may suggest ways
to improve holistic training. The results with \citeS{daume07} domain adaptation algorithm
(Section~\ref{subsec:decomp-holistic}) hint that holistic training may be able to serve
as a supplement to standard training.

Our main contribution is to extend the similarity composition approach \cite{turney12,turney13}
beyond recognition, to generation of composition and decomposition. By combining an unsupervised
first pass with a supervised second pass, we can scale up to the generation task.

\acks{Thanks to Charles Clarke and Egidio Terra for sharing the Waterloo corpus.
Thanks to my colleagues at NRC, who provided many helpful comments during a
presentation of this research. Thanks to Georgiana Dinu, Nghia The Pham, and Marco Baroni
for sharing their noun-adjective dataset and their vocabulary of 21,000 nouns.}

\vskip 0.2in

\bibliography{comp-decomp}
\bibliographystyle{theapa}

\end{document}